\documentclass{article}

\PassOptionsToPackage{numbers, compress}{natbib}
\newif\ifcameraready
\camerareadytrue
\ifcameraready
  \usepackage[main,preprint]{neurips_2026}
\else
  \usepackage[main]{neurips_2026}
\fi

\usepackage[utf8]{inputenc} 
\usepackage[T1]{fontenc}    
\usepackage{hyperref}       
\usepackage{url}            
\usepackage{booktabs}       
\usepackage{amsfonts}       
\usepackage{nicefrac}       
\usepackage{microtype}      
\usepackage{xcolor}         

\usepackage[ruled,vlined,linesnumbered]{algorithm2e}
\usepackage[normalem]{ulem}

\usepackage{amsmath,amssymb}
\usepackage{multirow}
\usepackage{array}
\usepackage{makecell}

\usepackage{tabularx}
\usepackage{graphicx}

\usepackage[nameinlink,noabbrev]{cleveref}

\usepackage{amsthm}

\newtheorem{remark}{Remark}
\newcommand{\softparagraph}[1]{\par\noindent\textit{\textbf{#1}}\ }

\usepackage{comment}

\title{Elastic Spectral State Space Models for Train-Once Budgeted Inference}

\ifcameraready
\author{%
  Dachuan Song$^{1}$,\quad Junyu Yin$^{2}$,\quad Zechen Hu$^{1}$,\quad Xuan Wang$^{1}$\thanks{Corresponding author: \texttt{xwang64@gmu.edu}.} \\
  $^{1}$Department of Electrical and Computer Engineering \\
  $^{2}$Department of Computer Science \\
  George Mason University, Fairfax, VA, USA
}
\else
\author{Anonymous Author(s)}
\fi

\begin{document}

\maketitle

\begin{abstract}
Modern sequence models are typically trained at a fixed computational capacity, while real-world applications require deployment across platforms with different resource constraints. 
Existing approaches either train or distill separate compact models for selected budgets, or use elastic architectures that shrink dimensions such as width, depth, Feed-Forward Network (FFN) size, or SSM state dimension.
In this paper, we propose Elastic Spectral State Space Models (ES-SSM), a train-once, export-many sequence modeling framework that gains elasticity through spectral approximation of the SSM sequence operator. 
ES-SSM builds on Hankel spectral filtering for state space models, where long-range token mixing is represented through fixed Hankel spectral channels that define an operator-level approximation resolution. 
To make this spectral elasticity reliably deployable, ES-SSM combines input-adaptive channel-wise gates with budget dropout, training the same spectral prefixes that are used by compact models at inference time.
This encourages low-index channels to become predictive on their own, while higher-index channels provide refinements for larger budgets. 
We evaluate ES-SSM across byte-level language modeling, Long Range Arena, Speech Commands V2, and offline reinforcement learning benchmarks. 
Across these settings, a single trained ES-SSM can be truncated to competitive compact models compared with modern Transformer and SSM baselines at similar parameter scales. Furthermore, by testing under various runtime budgets, we observe smooth and stable quality-cost curves over a wide range of truncation levels. Code is available at \url{https://github.com/GMU-MICO/ES-SSM}.


\end{abstract}




\section{Introduction}
\label{sec:intro}
Modern sequence models are typically trained at a fixed computational capacity. However, real-world applications require AI deployment across heterogeneous computational platforms, from high-throughput cloud clusters to compute-limited endpoints and power-constrained edge devices~\citep{bommasani2021opportunities}. 
This calls for models that can adapt to different trade-offs between inference cost and predictive performance~\citep{graves2016adaptive,teerapittayanon2016branchynet}. 
Common approaches address this challenge by training a family of model variants at different sizes~\citep{yu2018slimmable,cai2019once}, or by distilling a large teacher into one or more smaller students~\citep{hinton2015distilling,sanh2019distilbert,jiao2020tinybert,gou2021knowledge}. 
Although these methods can produce compact models for selected budgets, each supported capacity typically requires additional training, calibration, or fine-tuning.

Elastic neural networks provide a more flexible alternative by training a single model that supports multiple deployment budgets. 
Slimmable Networks and Once-for-All vary architectural choices such as width and depth~\citep{yu2018slimmable,cai2019once}, while Matryoshka representation learning, MatFormer, and MatMamba train nested representation dimensions, FFN widths, or SSM-dimension slices to be predictive~\citep{kusupati2022matryoshka,kudugunta2024matformer,shukla2024matmamba}. 
However, their elasticity is primarily architectural: smaller models are obtained by narrowing or removing parts of the network architecture. 
The choice of which architectural components to shrink is indirectly related to the final model quality, and therefore often requires careful design, training, or calibration.

In this paper, we study elasticity from a new operator-approximation perspective. 
For long-sequence modeling, especially in state space models (SSMs), the key computation is long-range token mixing, which can be represented by a convolutional sequence operator. 
This suggests an alternative route to elasticity: \textit{controlling the approximation resolution of the sequence operator}, rather than only shrinking architectural components.
Hankel spectral SSMs make this possible by expressing the operator using fixed spectral channels derived from a Hankel matrix~\citep{hazan2017learning,agarwal2023spectral}. 
Under this representation, elasticity is associated with a direct operator-level interpretation: using fewer spectral channels corresponds to a lower-resolution approximation of the sequence operator.
However, this spectral-resolution axis is not automatically deployable. 
A spectral SSM trained only at full capacity may distribute task-relevant information across many channels. As a result, naive channel truncation can discard important learned components and lead to sharp performance degradation.

\softparagraph{Statement of Contributions.}
To address the above issues, we propose Elastic Spectral State Space Models (ES-SSM), which train a full-capacity spectral SSM once and export compact models by direct prefix truncation along the Hankel spectral-channel axis. The main contributions are:

$\bullet$ \textbf{An elastic spectral SSM architecture.} 
    We introduce a spectral SSM sublayer built on fixed Hankel spectral channels, learned channel projections, and input-adaptive channel-wise gates. 
    The fixed spectral channels provide an ordered approximation axis for sequence mixing, while the learned projections and gates determine how strongly each channel contributes to the current representation. 
    This design preserves a prefix-structured channel axis, so compact models can be obtained by truncating later spectral channels rather than redesigning or reordering the model.

$\bullet$ \textbf{A budget-dropout training mechanism for reliable prefix truncation.} 
    We train ES-SSM under randomized spectral budgets that execute different retained prefixes of the spectral-channel axis during training. 
    As a result, the same prefix computations used by exported compact models are repeatedly optimized before deployment. 
    This training mechanism encourages early channels to carry useful predictive information independently, while later channels learn to improve the representation when larger budgets are available.

$\bullet$ \textbf{Train-once, export-many deployment and evaluation.} 
    For each runtime budget, ES-SSM exports a standalone compact model by keeping only the corresponding spectral-channel prefix and its associated parameters, without budget-specific retraining, channel search, or reordering.    
    We evaluate ES-SSM on byte-level SlimPajama against elastic and non-elastic Transformer and SSM baselines, and extend the same exported-compact-model protocol to Long Range Arena, Speech Commands V2, D4RL MuJoCo offline reinforcement learning, and a byte-level generality replication. 
    Across these settings, ES-SSM demonstrates superior or comparable performance relative to state-of-the-art elastic methods in quality--cost trade-off curves, sweet spots and collapse boundaries, and latency and memory measurements, while incurring substantially lower multi-budget training cost than independently trained non-elastic baselines.

\section{Related work}
\label{subsec:related_work}

\softparagraph{Budget-adaptive and elastic sequence models.}
A growing body of work studies how to adapt neural networks to different inference budgets. 
One class of methods produces compact models for selected deployment budgets. 
Knowledge distillation~\citep{hinton2015distilling,sanh2019distilbert,xu2024survey} transfers behavior from a larger teacher into one or more separately trained smaller students. Conditional computation~\citep{shazeer2017outrageously,fu2024lazyllm} changes the amount of active computation per input or token, for example by routing through a subset of modules or tokens, but typically does not export a standalone compact model with fewer retained parameters, so deployment may still require storing the full model. Structured pruning~\citep{flynn2024stat, wei2024structured} removes heads, channels, layers, or other parameter groups from a trained model. They often require additional selection criteria, calibration, or fine-tuning to recover quality.

Another line of work trains a single model that contains multiple nested subnetworks. 
Slimmable Networks~\citep{yu2018slimmable} train a single network that supports multiple width multipliers, with shared weights and switchable normalization so that each width can be evaluated as a valid subnetwork.
Once-for-All~\citep{cai2019once} extends this idea to a larger architectural search space, where subnetworks at different depths, widths, kernel sizes, or resolutions are inherited from one trained supernet.
Structured dropout over Transformer layers enables layer-pruned shallower subnetworks to be used at inference time~\citep{fan2019reducing}. 
Matryoshka Representation Learning introduces nested feature-vector prefixes, where each prefix is supervised to be predictive.
MatFormer~\citep{kudugunta2024matformer} adapts this nesting principle to Transformer language models through nested FFN blocks, while MatMamba~\citep{shukla2024matmamba} introduces nested capacity along Mamba-2-style SSM dimensions.
These elastic methods primarily vary architectural axes such as width, depth, FFN size, representation dimension, or SSM hidden dimension. 
In contrast, the proposed ES-SSM achieves elasticity through the Hankel spectral channels of the SSM sequence operator. 
This operator-level elasticity is complementary to architectural nesting. 

\softparagraph{Hankel spectral filtering and spectral SSMs.}
In addition to attention-based modeling, state space models (SSMs) have
re-emerged as an efficient alternative for long-sequence modeling. SSMs view
sequence processing as a dynamical system that repeatedly compresses new
inputs into latent states, which can serve as a stable long-range memory
mechanism~\citep{gu2020hippo,gu2021combining}. Subsequent work has improved
the trainability, numerical robustness, and computational efficiency of SSMs
through structured parameterizations, fast convolution, and scan-based
implementations~\citep{gu2021efficiently,gupta2022diagonal,song2024causality,smith2022simplified}.
These developments show that carefully parameterized linear recurrent layers
can achieve strong long-context performance with favorable scaling in sequence
length compared with attention-based models~\citep{orvieto2023resurrecting}.
However, in most existing formulations, the training and deployment compute
budgets are fixed by architectural choices such as the number of channels or
state dimensions. Reducing inference-time cost therefore typically requires
training a smaller model variant or applying a separate compression or pruning
procedure.

Building on SSMs, a distinct line of research constructs sequence operators
from Hankel structure and spectral filtering. It reformulates SSM channels in
the spectral domain while retaining provable learning guarantees in dynamical
settings~\citep{hazan2017learning,hazan2018spectral}. Recent spectral SSMs
build on this perspective by fixing a Hankel spectral eigenbasis and learning
a small set of mixing parameters over precomputed filters, enabling efficient
long-range sequence computation~\citep{agarwal2023spectral}. Because the
basis is fixed and ordered, this formulation induces a natural notion of
spectral resolution: the sequence operator can be represented using a prefix
subset of spectral components.

However, prior work typically trains and evaluates models at a single fixed
spectral resolution, treating truncation mainly as a post hoc approximation.
In contrast, our deployment setting requires truncation to remain reliable
when all runtime budgets are derived from one trained set of parameters. 

\section{Preliminaries}
\label{sec:preliminaries}

\softparagraph{Setup and notation.} Let $u_{1:L}=(u(1),\ldots,u(L))$ denote an input sequence of length $L$, with $u(t)\in\mathbb{R}^{d_{\mathrm{in}}}$, and let $ y_{1:L}=( y(1),\ldots, y(L))$ denote the layer output, with $ y(t)\in\mathbb{R}^{d_{\mathrm{out}}}$. Here $d_{\mathrm{in}}$ and $d_{\mathrm{out}}$ are the input and output feature dimensions, respectively. In residual model blocks where the sublayer preserves the representation width, we set $d_{\mathrm{in}}=d_{\mathrm{out}}=d$ and refer to $d$ as the model dimension. Whenever an expression refers to indices $t\le 0$, we assume zero-padding. All inner products $\langle\cdot,\cdot\rangle$ are the standard Euclidean inner product, i.e.\ $\langle v,w\rangle = v^{\top}w$.


\softparagraph{Linear dynamical systems and Hankel spectral basis.}
A (discrete-time) state space system (SSM) evolves as
$x(t) = A\,x(t-1) + B\,u(t),$
and
$y(t) = C\,x(t) + D\,u(t),$
with latent state $x(t)\in\mathbb{R}^{N}$, input $u(t)\in\mathbb{R}^{d_{\mathrm{in}}}$,
output $y(t)\in\mathbb{R}^{d_{\mathrm{out}}}$, and matrices
$A\in\mathbb{R}^{N\times N}$,
$B\in\mathbb{R}^{N\times d_{\mathrm{in}}}$,
$C\in\mathbb{R}^{d_{\mathrm{out}}\times N}$,
$D\in\mathbb{R}^{d_{\mathrm{out}}\times d_{\mathrm{in}}}$.
Assuming zero initial state, the input--output map of
SSM admits a causal convolution:
\begin{equation}
y(t)
= D\,u(t) + \sum_{\tau=0}^{t-1} G(\tau)\,u(t-\tau)
= D\,u(t) + (\mathbf{G}*u)(t),
\qquad t=1,\ldots,L,
\label{eq:conv_view}
\end{equation}
where $G(\tau)=CA^{\tau}B\in\mathbb{R}^{d_{\mathrm{out}}\times d_{\mathrm{in}}}$
is the \emph{causal kernel} and $\mathbf{G}=\{G(\tau)\}_{\tau\ge 0}$.

\softparagraph{Spectral parameterization of SSM.}
We first introduce the base spectral parameterization used in this work.
Consider the length-$L$ causal convolutional kernel
$\mathbf{G}=\{G(\tau)\}_{\tau=0}^{L-1}$ induced by a stable linear state-space
model. We follow the standard stable symmetric setting used in spectral
filtering analyses \citep{hazan2017learning,agarwal2023spectral}, where
$A$ is real symmetric with eigenvalues in $[0,1)$. Thus $A$ admits the
orthogonal eigendecomposition $A=U\Lambda U^\top$, with
$\Lambda=\mathrm{diag}(\lambda_1,\ldots,\lambda_N)$ and
$\lambda_\ell\in[0,1)$. The causal kernel can therefore be written in
diagonal coordinates as
\begin{equation}
G(\tau)
=
C A^\tau B
=
C U \Lambda^\tau U^\top B
=
\sum_{\ell=1}^{N}
\lambda_\ell^\tau\,
(CU)_{:\ell}\,(U^\top B)_{\ell:},
\label{eq:G_diag_decomp}
\end{equation}
where $(CU)_{:\ell}\in\mathbb{R}^{d_{\mathrm{out}}}$ denotes the $\ell$-th
column of $CU$, and $(U^\top B)_{\ell:}\in\mathbb{R}^{1\times d_{\mathrm{in}}}$
denotes the $\ell$-th row of $U^\top B$. Thus, the matrix-valued kernel
decomposes into a sum of scalar temporal traces and input-output coefficient
matrices. The only time-dependent factor in each term is the geometric
sequence $[1,\lambda_\ell,\lambda_\ell^2,\ldots,\lambda_\ell^{L-1}]$.
Therefore, approximating the SSM kernel over a finite horizon reduces to
approximating geometric vectors
$\nu(\lambda)=[1,\lambda,\lambda^2,\ldots,\lambda^{L-1}]^\top
\in\mathbb{R}^{L}$, with $\lambda\in[0,1)$.

Spectral filtering provides a universal way to approximate these geometric
vectors. Specifically, one constructs a set of Hankel spectral channels
$[\phi_1,\phi_2,\ldots,\phi_L\in\mathbb{R}^{L}],$
which depend only on the sequence length \(L\), not on the hidden dimension
\(N\) or on a particular realization of \(A,B,C\). 
We can project the geometric vector $\nu(\lambda)$ onto the span of the first
$K$ Hankel spectral channels $\{\phi_k\}_{k=1}^{K}$ to obtain the low-rank
approximation $\nu(\lambda)
  \approx
   \sum_{k=1}^{K}
   \langle \nu(\lambda),\phi_k\rangle \phi_k$.
   For each $\lambda^\tau$ in $\nu(\lambda)$, it follows $\lambda^\tau \approx\sum_{k=1}^{K}
\langle \nu(\lambda),\phi_k\rangle \phi_k(\tau),$
where \(\phi_k(\tau)\) denotes the \((\tau+1)\)-th entry of the channel vector
\(\phi_k\).
Substituting this approximation into the causal kernel yields
\[
G(\tau)=\sum_{\ell=1}^{N}
\lambda_\ell^\tau\,
(CU)_{:\ell}(U^\top B)_{\ell:}
\approx
\sum_{\ell=1}^{N}
\sum_{k=1}^{K}
\langle\nu(\lambda_\ell),\phi_k\rangle\,
\phi_k(\tau)\,
(CU)_{:\ell}(U^\top B)_{\ell:}=\sum_{k=1}^{K} M_k\phi_k(\tau)
\]
for $\tau=0,\ldots,L-1$,
where each $M_k=
\sum_{\ell=1}^{N}
\langle\nu(\lambda_\ell),\phi_k\rangle\,
(CU)_{:\ell}(U^\top B)_{\ell:}\in\mathbb{R}^{d_{\mathrm{out}}\times d_{\mathrm{in}}}$
is unknown and
collects the input-output coefficient matrices associated with the $k$-th
spectral channel. The detailed derivation of $M_k$ and the Hankel channel
construction are provided in Appendix~\ref{app:spectral_derivation}.

Let
    $\Phi_k=\{\phi_k(\tau)\}_{\tau=0}^{L-1}$
be a fixed causal convolution filter, the corresponding spectral SSM becomes: 
\begin{equation}
    \hat y(t)
    =
    \sum_{k=1}^{K} M_k(\Phi_k*u)(t),
    \label{eq:base_spectral_ssm}
\end{equation}
where \(*\) denotes causal convolution. The direct feedthrough term $D\,u(t)$ in \cref{eq:conv_view} is a zero-lag causal contribution. Since our convolution convention includes the $\tau=0$ term, we absorb it into the learned kernel by setting $G(0)\leftarrow G(0)+D$ and keep the notation $\mathbf{G}$ for the absorbed kernel.

\softparagraph{Limitations of the base spectral form.}
\label{subsec:base_spectral_limitations}
The approximation in \cref{eq:base_spectral_ssm} suggests a natural route to
elastic SSM deployment through direct prefix truncation. In principle, one could train a model once at a full spectral budget $\overline K$ and, for a smaller runtime budget $K\le\overline K$, export a compact model that keeps spectral channels $1,\ldots,K$ and removes channels $k>K$.
However, direct prefix truncation exposes a key limitation of the base
spectral form:
Hankel eigenvalue order gives an approximation order for
the basis, but full-budget training of \cref{eq:base_spectral_ssm} does
not make this order, in terms of information importance, for the learned channel
projections $\{M_k\}$.  Since the training loss is applied only to the aggregate output
$\sum_{k=1}^{\overline K}M_k(\Phi_k*u)(t)$, predictive information can be distributed across all channels. In particular, important information may be placed in high-index channels, or encoded through cancellations and compensations between low-index and high-index channels. 
Consequently, simply removing channels \(k>K\) after training may discard
essential learned components rather than channels that only serve as refinements.
This can cause sharp degradation at small budgets.

\section{Methods}
\label{sec:Methods}
We aim to train a single full-capacity ES-SSM with maximum spectral budget
$\overline K$ and export compact models for smaller runtime budgets
$K\le\overline K$. As discussed in \cref{subsec:base_spectral_limitations},
direct prefix truncation is useful only if the learned use of Hankel spectral
channels respects the prefix order: low-index channels should already form a
useful predictor, while later channels should provide higher-capacity
refinements. ES-SSM enforces this through two coupled mechanisms:
First, it adds input-adaptive, channel-wise sigmoid gates so that each retained
Hankel spectral channel can be modulated according to the current
representation. Second, it uses budget dropout, where each update trains the
same prefix computation that will later be used by an exported compact model.
Together, these mechanisms learn one ordered set of channel projections
$\{M_k\}_{k=1}^{\overline K}$, rather than separate projection sets for each
budget.

\begin{figure*}[t]
\centering
\includegraphics[width=0.9\textwidth]{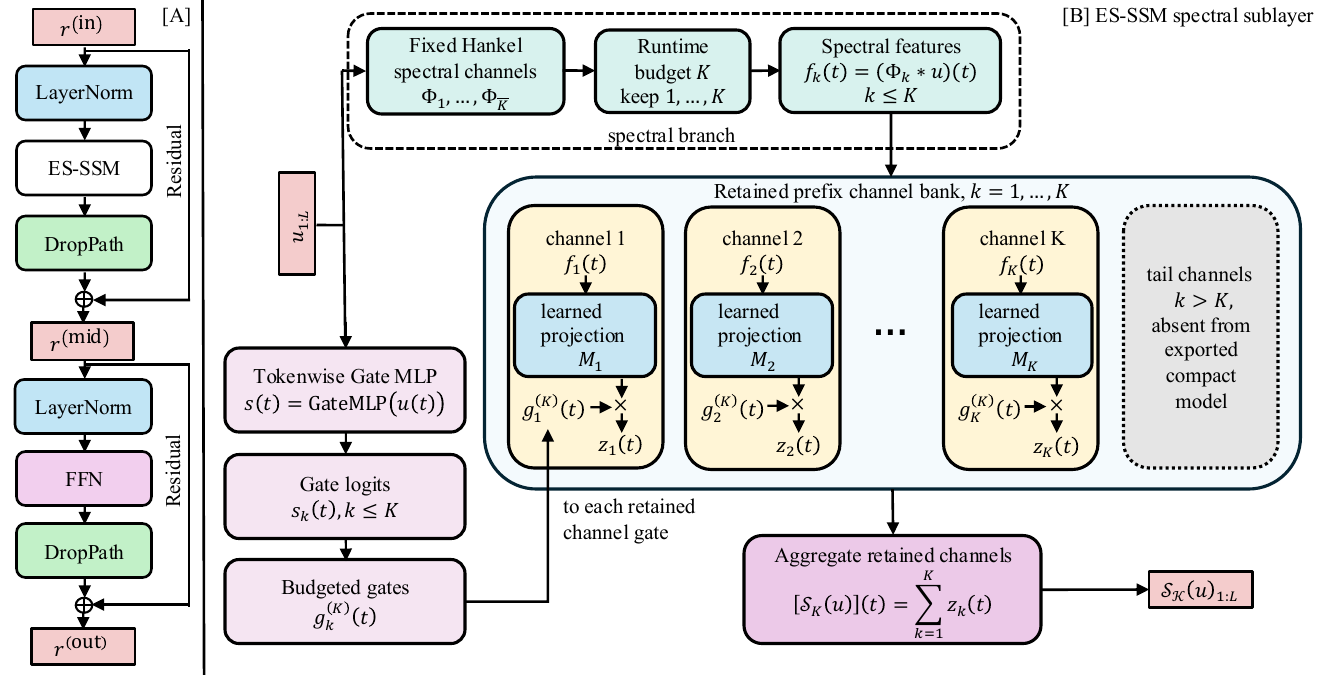}
\caption{ES-SSM architecture overview.
[A] ES-SSM is embedded in a pre-norm residual block with an ES-SSM spectral sublayer followed by an FFN; residual paths carry the block input around each sublayer.
[B] The ES-SSM spectral sublayer $\mathcal{S}_K$ combines fixed Hankel spectral channels with a tokenwise Gate MLP that produces input-adaptive, channel-wise budgeted gates. For a runtime budget $K\le \overline K$, direct prefix truncation keeps channels $1,\ldots,K$ in the exported compact model and removes tail channels $k>K$. Each retained channel is projected by $M_k$, modulated by the scalar gate $g_k^{(K)}(t)$, and added to the retained-channel sum. Thus, one full-capacity training can be truncated into compact models at multiple runtime budgets.
}
\label{fig:model_overview}
\end{figure*}

At deployment time, for a runtime budget $K$, we apply direct prefix truncation to keep spectral channels
$1,\ldots,K$ and their channel-indexed parameters, and remove channels
$k>K$. The exported compact model is therefore physically smaller along the
spectral-channel axis and requires no budget-specific retraining, channel search,
or reordering. \Cref{fig:model_overview} summarizes the pre-norm residual block
and the ES-SSM spectral sublayer used inside it. 
Building on the base spectral
form in \cref{eq:base_spectral_ssm}, ES-SSM assigns each retained Hankel
spectral channel a learned projection $M_k$ and an input-adaptive scalar gate.
The projection provides budget-shared channel mixing, while the gate gives the
sublayer a tokenwise mechanism to suppress currently irrelevant channels and
activate useful spectral refinements. Budget dropout then trains these gated
prefixes to remain predictive across runtime budgets. We denote the logistic
sigmoid by $\operatorname{sigmoid}(a)=1/(1+\exp(-a))$.

\subsection{Spectral sublayer}
\label{subsec:spectral_sublayer}
For a sublayer input sequence $u_{1:L}$, the gate logits are produced
tokenwise by a lightweight two-layer MLP with GELU activation,
$\mathbf{s}(t)=
\mathbf{W}_{g,2}\mathrm{GELU}(\mathbf{W}_{g,1}u(t)+\mathbf{b}_{g,1})
+\mathbf{b}_{g,2}$, where
$\mathbf{s}(t)=[s_1(t),\ldots,s_{\overline K}(t)]^\top
\in\mathbb{R}^{\overline K}.$
The parameters are
$\mathbf{W}_{g,1}\in\mathbb{R}^{d_g\times d_{\mathrm{in}}}$,
$\mathbf{b}_{g,1}\in\mathbb{R}^{d_g}$,
$\mathbf{W}_{g,2}\in\mathbb{R}^{\overline K\times d_g}$, and
$\mathbf{b}_{g,2}\in\mathbb{R}^{\overline K}$, with hidden width
$d_g=d_{\mathrm{in}}/2$. We initialize
$\mathbf{b}_{g,2}\equiv -2$, so each channel starts with a small gate
value, $\operatorname{sigmoid}(-2)\approx 0.12$. This keeps the spectral
residual contribution small at initialization and avoids large random
channel sums before budget dropout has trained useful prefixes.

For a runtime budget $K\le\overline K$, the exported compact model keeps
the first $K$ logits. The applied gate coefficient for channel $k\le K$ is
$g^{(K)}_k(t)=\frac{1}{\sqrt{K}}\operatorname{sigmoid}(s_k(t))$, where the
factor $1/\sqrt{K}$ is intended to keep the accumulated spectral contribution
on a comparable scale across budgets.
Let $\mathcal{S}_K$ denote the ES-SSM spectral sublayer at runtime budget
$K$. Its output sequence $\mathcal{S}_K(u)_{1:L}$ is defined entrywise by
\begin{equation}
[\mathcal{S}_K(u)](t)
=
\sum_{k=1}^{K}
g^{(K)}_k(t)\;M_k\,(\Phi_k*u)(t),
\label{eq:adaptive_spectral_K}
\end{equation}
In \cref{fig:model_overview}, we abbreviate each retained-channel contribution as
$z_k(t)\triangleq g_k^{(K)}(t)M_k(\Phi_k*u)(t)$. Here $\Phi_k$ is the fixed
Hankel spectral channel viewed as a causal kernel sequence, and
$M_k\in\mathbb{R}^{d_{\mathrm{out}}\times d_{\mathrm{in}}}$ is the
corresponding learned channel projection.
Compared with \cref{eq:base_spectral_ssm}, the sublayer adds input-dependent
channel modulation and the budget scale $1/\sqrt{K}$, while retaining the same prefix-ordered channel structure. 

The spectral sublayer is embedded in a pre-norm residual block. Let
$r^{(\mathrm{in})}_{1:L}$, $r^{(\mathrm{mid})}_{1:L}$, and
$r^{(\mathrm{out})}_{1:L}$ denote the block input, intermediate sequence,
and block output.
LayerNorms $\mathrm{LN}_1,\mathrm{LN}_2$ are applied tokenwise, $\mathrm{DropPath}$ denotes stochastic depth on the residual branch, and $\mathrm{FFN}$ is a tokenwise two-layer MLP with widening factor $4$ and GELU activation. The block updates as
$$r^{(\mathrm{mid})}
=
r^{(\mathrm{in})}
+
\mathrm{DropPath}\!\bigl(
\mathcal{S}_K(\mathrm{LN}_1(r^{(\mathrm{in})}))
\bigr),~~r^{(\mathrm{out})}
=
r^{(\mathrm{mid})}
+
\mathrm{DropPath}\!\bigl(
\mathrm{FFN}(\mathrm{LN}_2(r^{(\mathrm{mid})}))
\bigr)$$


The $K$ spectral convolutions $\{(\Phi_k*u)(t)\}_{k=1}^{K}$ can be computed
jointly by a single batched FFT. Direct prefix truncation slices every
channel-indexed tensor along the spectral-channel axis, including the first
$K$ spectral channels, the first $K$ rows of $\mathbf{W}_{g,2}$, the first
$K$ entries of $\mathbf{b}_{g,2}$, and the projection stack $\{M_k\}_{k=1}^K$.
Implementation details on precision, batching, and gate-before-projection
ordering are given in Appendix~\ref{app:efficient_impl}.

\softparagraph{Budgeted inference with exported compact models.}
At deployment time, ES-SSM exports a compact model for runtime budget $K$ by
direct prefix truncation. The exported compact model keeps spectral channels
$1,\ldots,K$ and removes channels $k>K$. Equivalently, it keeps
$\{\Phi_k,M_k\}_{k=1}^{K}$, the first $K$ rows of $\mathbf{W}_{g,2}$, and the
first $K$ entries of $\mathbf{b}_{g,2}$, while retaining shared parameters such
as the embedding, $\mathbf{W}_{g,1}$, $\mathbf{b}_{g,1}$, LayerNorms, FFN, and
output head. Because retained gate logits $s_k(t)$ for $k\le K$ do not depend
on later channels, the export is mechanically prefix-consistent up to the
deterministic scale $1/\sqrt{K}$. Thus, deployment requires no channel search or
budget-specific reordering. Next, we introduce budget dropout, which facilitates the above model structure by training the exported prefixes to be effective predictors.
The exported-inference procedure is given as Algorithm~\ref{alg:inference} in
Appendix~\ref{app:method_algorithms}.

\subsection{Budget dropout training}
\label{subsec:budget_adaptive_training}

Budget dropout trains the same prefixes that will later be exported for
budgeted inference. Let $F_{\theta}(u;K)$ denote the full ES-SSM network
evaluated with every spectral sublayer using runtime budget $K$. By default, at each optimization step, we choose a training budget
$K_{\mathrm{train}}\le \overline K$ and minimize
$\mathcal{L}_{\mathrm{train}}(\theta)=
\mathcal{L}_{\mathrm{task}}(F_{\theta}(u;K_{\mathrm{train}}),y)$.
Thus, the forward pass during training matches the direct prefix computation
used by the exported compact model.

Budget dropout changes the gradient allocation across spectral channels. At budget
$K_{\mathrm{train}}$, only channels $1,\ldots,K_{\mathrm{train}}$ participate
in the spectral computation. For channels $k>K_{\mathrm{train}}$, the channel-specific trainable
parameters, including $M_k$, row $k$ of $\mathbf{W}_{g,2}$, and entry $k$ of
$\mathbf{b}_{g,2}$, receive no gradient on that step. In contrast, shared
parameters such as embeddings, $\mathbf{W}_{g,1}$, $\mathbf{b}_{g,1}$,
LayerNorms, FFN, and the output head are updated whenever they are used by the
full network.
Since a low-index channel appears in more sampled prefixes than a high-index
channel, budget dropout aggregates information towards low-index channels, making them effective for many exported compact models (i.e., different $K$ values). On the other hand, higher-index channels only act as refinements.

\begin{figure*}[t]
\centering
\includegraphics[width=\textwidth]{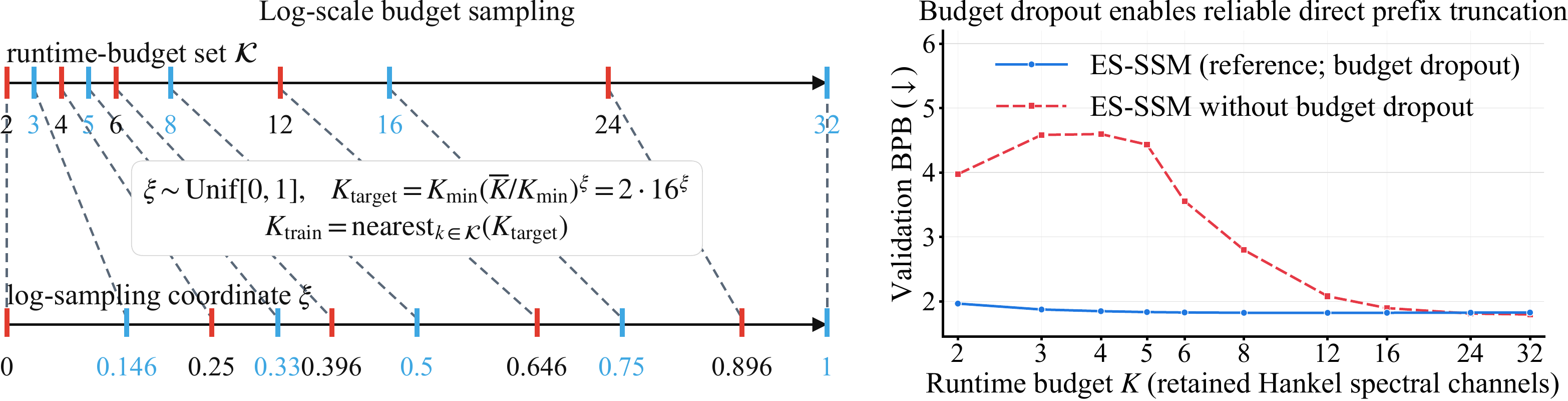}
\caption{Budget dropout sampler and prefix-truncation behavior.
Left: ES-SSM samples a log-scale target $K_{\mathrm{target}}$ and snaps it to the nearest element of the runtime-budget set $\mathcal{K}$, placing more training pressure on small-$K$ prefixes while preserving coverage of larger budgets. Right: budget dropout trains the exported prefixes directly, yielding stable validation bits-per-byte (BPB) across runtime budgets; without budget dropout, a model trained only at $\overline K$ degrades sharply under direct prefix truncation.}
\label{fig:budget_dropout}
\end{figure*}

We sample $K_{\mathrm{train}}$ from a runtime-budget set
$\mathcal{K}$ and set $K_{\min}=\min\mathcal{K}$.
The training involves three steps. First, during the first $E_{\mathrm{warmup}}$
epochs, we set $K_{\mathrm{train}}=\overline K$ so that all channel-specific
parameters are initialized under the full computation path. 
Second, after warmup, every $n_{\mathrm{full}}$ optimization steps we include a full-budget
anchor update with $K_{\mathrm{train}}=\overline K$; this ensures that tail
channels (at least) periodically receive gradients to maintain their effectiveness. Third, for
all remaining steps, we draw a log-scale $\xi\sim\mathrm{Unif}[0,1]$ (cf. \cref{fig:budget_dropout}) between $K_{\min}$ and $\overline K$, and choose the nearest element in
$\mathcal{K}$:
\begin{equation}
K_{\mathrm{train}}
=
\underset{k\in\mathcal{K}}{\arg\min}\;
\Big|k-K_{\min}\,(\overline K/K_{\min})^{\xi}\Big|.
\label{eq:budget_sampler}
\end{equation}
Log-scale sampling treats budgets as multiplicative capacity levels rather than
absolute channel counts. It increases training pressure on small-$K$ prefixes
where each channel has larger marginal effect, while still covering large
budgets through sampling and full-budget anchor updates.

We use $n_{\mathrm{full}}=8$, giving one full-budget anchor update every eight
post-warmup steps; Appendix~\ref{app:budget_sampler} and
Appendix~\ref{app:abl_full_every} explain this setting and analyze the anchor
cadence. The budget-dropout training procedure is given as Algorithm~\ref{alg:training} in Appendix~\ref{app:method_algorithms}. In our sensitivity experiments, validation BPB changes little when sampling all integer budgets or removing the full-budget warmup, as detailed in Appendices~\ref{app:runtime_budget_set_sensitivity} and~\ref{app:warmup_sensitivity}.

\section{Experiments}
\label{sec:experiments}

The empirical study has two goals. The first is to verify that, at modern foundation-model scale, training a single ES-SSM at full capacity $\overline K=32$ produces a family of exported compact models whose per-budget quality can approach or exceed independently trained sequence models at the same parameter counts. The second is to verify that the budgeted-inference claim transfers across modalities, not just byte-level language modeling, but also long-range classification, audio, and offline decision-making.

ES-SSM is trained once at the maximum spectral budget $\overline K=32$ and exported to smaller runtime budgets $K\in\{2,3,4,5,6,8,12,16,24,32\}$ by direct prefix truncation (\cref{sec:Methods}). For ES-SSM, $K$ has its architectural meaning as the runtime spectral-prefix budget. For baselines, $K$ is only a matched-budget label: each point uses a model with parameter count comparable to ES-SSM at the corresponding budget. Within each benchmark suite, we match the optimizer, schedule, batch size, precision policy, and evaluation protocol across model families so that the comparison isolates architecture. Main tables report both task quality (bits-per-byte (BPB)\footnote{For a byte sequence $x_{1:T}$, $\mathrm{BPB}=-(T\log 2)^{-1}\sum_{t=1}^{T}\log p_{\theta}(x_t\mid x_{<t})$.} or task-appropriate accuracy / return) and inference cost (latency, throughput, peak memory) on the same hardware; full hyperparameters and parameter-matching details are 
provided in
Appendix~\ref{app:setup}.

\softparagraph{Training-cost accounting.} All ES-SSM budget points come from one training run, whereas non-elastic Transformer and SSM baselines require one training run per budget. Thus, ES-SSM pays training cost once and amortizes it across deployment budgets. In a controlled comparison with always-full training, budget dropout reaches matched full-budget validation quality with $17.3\%$ more optimization steps but $30.2\%$ less analytic compute and $39.4\%$ less wall-clock time, as detailed in Appendix~\ref{app:matched_quality_training_cost}.

\subsection{Byte-level language modeling at scale}
\label{subsec:exp_slimpajama}

\softparagraph{Setup.} We train all models on the SlimPajama-627B byte stream~\citep{sobolevaslimpajama} at sequence length $L=2048$ with vocabulary size $258$ ($256$ bytes plus PAD and BOS). The main configuration trains each run for $30$\,B byte-tokens and uses the measured per-budget parameter counts reported in Appendix~\ref{app:slimpajama_full}, Table~\ref{tab:slimpajama_main}; the ES-SSM full-budget checkpoint has around $1.3$\,B parameters. We compare ES-SSM against two prior elastic models, MatFormer~\citep{kudugunta2024matformer} and MatMamba~\citep{shukla2024matmamba}, and three non-elastic backbones, Transformer++~\citep{touvron2023llama}, Pythia-style~\citep{biderman2023pythia}, and GLA~\citep{yang2023gated}. The elastic families train once and export all ten runtime budgets, while the non-elastic baselines are trained independently at four reference budgets $K\in\{2,8,16,32\}$. All runs use one NVIDIA B200 GPU. We omit Mamba-2 from the training suite.

\begin{figure*}[t]
\centering
\includegraphics[width=\textwidth]{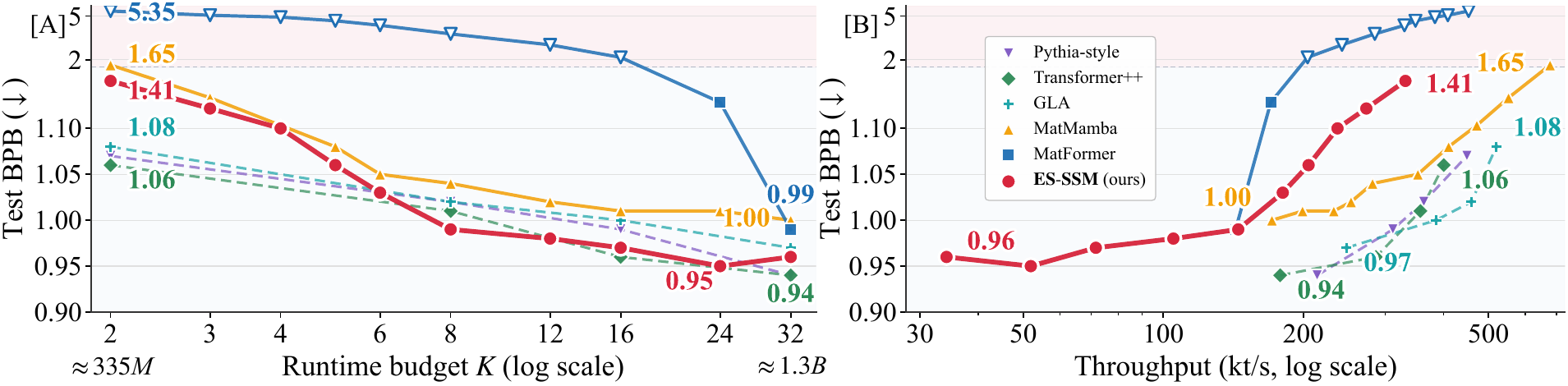}
\caption{SlimPajama main-table visualization for the $1.3$B-family, $30$B-token scaling experiment.
[A] Test BPB (lower is better) vs.\ runtime budget $K$ (log scale).
[B] Test BPB vs.\ inference throughput (kt/s, log scale;
sequence length $L{=}2048$, batch $1$, single B200 GPU).
The two panels share the same legend on the right.
Solid lines denote methods with elastic capabilities, whereas dashed lines denote non-elastic baselines trained \textbf{independently} at each shown capacity. We note that these dashed non-elastic baselines require significantly more training time and therefore, should \textbf{not} be used for direct comparison with our method. Instead, they are presented as oracle results for reference.
Markers are the data points in Appendix~\ref{app:slimpajama_full}, Table~\ref{tab:slimpajama_main}; for ES-SSM, the ten $K$ values instantiate exported parameter budgets spanning roughly $330$\,M to around $1.3$\,B.
The light-pink band highlights the high-BPB degradation region (Test BPB $>1.5$).
Inline values annotate the headline operating points.
For readability, both panels use a piecewise-linear $y$-axis that stretches
the competitive band $[0.90,1.10]$ and compresses the collapse range.
Per-family numerical results are reported in Appendix~\ref{app:slimpajama_full}.}
\label{fig:slimpajama_results}
\end{figure*}

\softparagraph{Main comparison.} \cref{fig:slimpajama_results} visualizes the main SlimPajama table from two complementary views: quality as a function of the runtime spectral budget $K$, and quality as a function of measured inference throughput. Elastic families are evaluated at all ten runtime budgets, whereas the non-elastic families contribute four independently trained, parameter-matched reference points. Detailed numerical results are reported in Appendix~\ref{app:slimpajama_full}, Table~\ref{tab:slimpajama_main}; full per-benchmark tables are deferred to Appendix~\ref{app:add_results}.

\softparagraph{Main result analysis.} \Cref{fig:slimpajama_results} and \cref{tab:slimpajama_main} show that ES-SSM produces a smooth, usable budget curve from one full-capacity training run. In \cref{fig:slimpajama_results}A, increasing $K$ steadily improves ES-SSM test BPB, with the curve moving from the low-budget regime toward the competitive band by moderate budgets and reaching its best operating region at large budgets. Importantly, even the smallest ES-SSM prefixes remain outside the collapse zone, while MatFormer suffers severe low-budget degradation and MatMamba remains on a higher-BPB trajectory before approaching its best value only at large $K$. \Cref{fig:slimpajama_results}B shows the same behavior from the deployment perspective: reducing $K$ moves ES-SSM toward substantially higher throughput while preserving a controlled quality degradation, so the spectral budget acts as a practical runtime knob rather than only a parameter-count index. At moderate and large budgets, ES-SSM remains competitive with independently trained Transformer++ / Pythia-style / GLA reference points, while avoiding the severe low-budget collapse seen in MatFormer and the higher-BPB plateau of MatMamba. The main reason is that ES-SSM makes the elastic axis the Hankel spectral-channel prefix: budget dropout trains the same prefixes used at inference, and the input-adaptive gates let each retained channel contribute conditionally rather than acting as a fixed spectral mixture. Thus the reported ES-SSM points should be read as exported compact models from one checkpoint, not as separately trained models for each budget.

\subsection{Cross-domain validation}
\label{subsec:exp_xdomain}

\begin{figure*}[t]
\centering
\includegraphics[width=\textwidth]{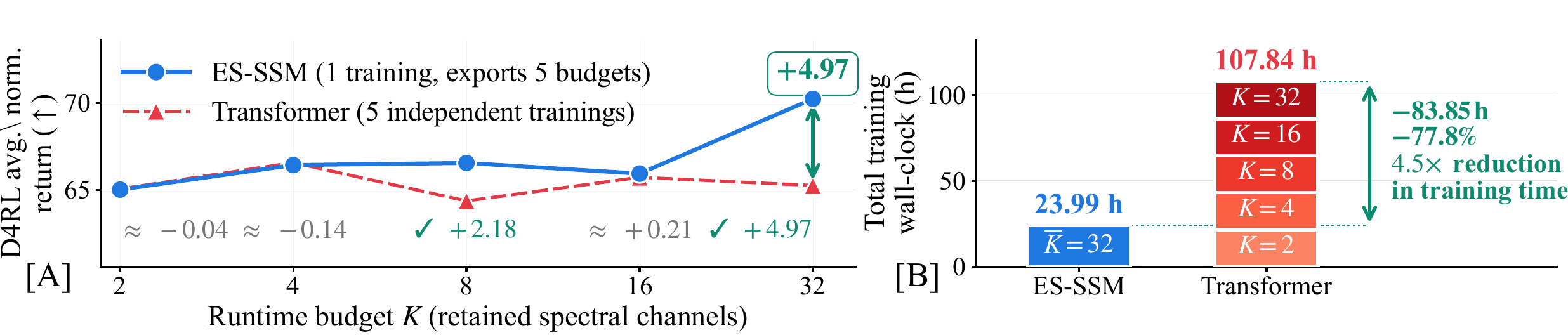}
\caption{
D4RL quality and training-cost comparison.
[A] Quality (avg.\ normalized return, larger means better) vs.\ runtime budget $K$.
ES-SSM (solid blue) is a single $\overline{K}{=}32$ training exported
at five budgets via prefix truncation; Transformer (dashed red) is
five \emph{independently trained} models, with the dashed connector
serving only as a visual guide. The bottom strip shows the per-$K$
delta (ES-SSM $-$ Transformer); $\checkmark$, $\approx$, $\times$
denote win, tie, and loss (tie threshold $0.5$\,pts).
[B] Total training wall-clock on the full D4RL suite
(9 datasets $\times$ 3 seeds, RTX~3090). The Transformer bar is
segmented into its five separate training runs, while ES-SSM trains
a single model. The teal arrow reports the savings. Full numerical tables are
reported in Appendix~\ref{app:add_results_d4rl}.
}
\label{fig:d4rl_value}
\end{figure*}

\begin{figure*}[t]
\centering
\includegraphics[width=\textwidth]{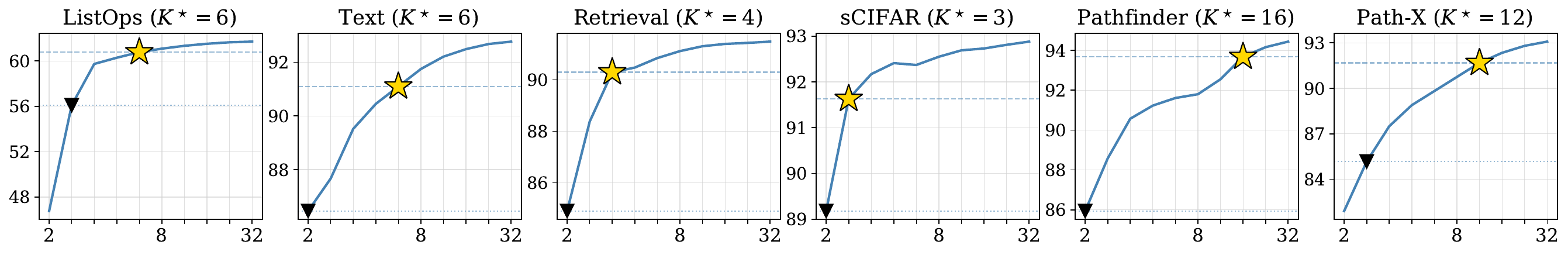}
\caption{LRA budget sweep at the matched $\sim\!50$\,M scale. Each curve is obtained by direct prefix truncation from the same $\overline K{=}32$ ES-SSM checkpoint. Gold stars mark the smallest $K$ retaining at least $98\%$ of the full-budget score; black triangles mark the smallest tested budget $K$ that retains at least $90\%$ of the full-budget score. Across all six tasks, ES-SSM degrades smoothly as $K$ decreases rather than collapsing abruptly. Full LRA tables are reported in Appendix~\ref{app:add_results_lra}.}
\label{fig:lra_budget_sweep}
\end{figure*}

To test whether budgeted inference is specific to byte-level language modeling or persists more broadly, we evaluate ES-SSM on four additional settings: long-range classification, audio, offline decision-making, and a byte-level generality replication at around $330$\,M parameters / $10$\,B byte-tokens. 
\Cref{fig:d4rl_value} highlights the quality and training-cost trade-off on offline decision-making, while \cref{fig:lra_budget_sweep} visualizes the LRA budget sweep. Detailed LRA, Speech Commands, D4RL, and byte-level generality tables are reported in Appendix~\ref{app:add_results_lra}, Appendix~\ref{app:add_results_sc10}, Appendix~\ref{app:add_results_d4rl}, and Appendix~\ref{app:add_results_byte_general}, respectively.


\softparagraph{Cross-domain analysis.} Across these additional domains, ES-SSM remains competitive while preserving the same deployment rule: low-budget points are exported compact models from a single full-capacity checkpoint, not separately trained models. On LRA, ES-SSM gives the strongest \textsc{Path-X} result in \cref{fig:lra_budget_sweep} and reaches the $98\%$ sweet spot at $K{=}12$; the full sweep shows gradual degradation rather than abrupt collapse. On Speech Commands V2, ES-SSM is within $0.27$ percentage points of Conformer at full capacity, while $K{=}3$ already retains almost all of the accuracy. On D4RL, \cref{fig:d4rl_value}A shows that ES-SSM is not merely cheaper to deploy: its exported budgets match the independently trained Transformer models at $K{=}2,4,16$ and improve the average normalized return at $K{=}8$ and $K{=}32$, with the largest gain appearing at full capacity. This pattern suggests that the spectral prefix gives useful compact policies at small budgets, while the additional channels at larger budgets provide meaningful refinement for offline control. \Cref{fig:d4rl_value}B further separates model quality from training cost: ES-SSM obtains all five D4RL deployment budgets from one training run, reducing total D4RL training wall-clock from $107.84$ h to $23.99$ h, a saving of $83.85$ h ($77.8\%$) and $4.5{\times}$ reduction in total training time. The byte-level generality replication is a sanity check for the training recipe: ES-SSM reaches the best validation BPB among the three compared architectures at that scale. Together, these results support the central claim that the advantage is not an isolated single-budget win, but a deployable family of compact sequence models obtained by direct prefix truncation.


\section{Conclusion}

We introduced ES-SSM, a train-once, export-many spectral SSM framework for budgeted sequence modeling. By combining input-adaptive channel-wise gates with budget dropout, ES-SSM makes direct prefix truncation reliable along the Hankel spectral-channel axis, allowing one full-capacity checkpoint to export compact models across runtime budgets. Across byte-level language modeling, LRA, Speech Commands V2, D4RL, and byte-level generality checks, ES-SSM yields smooth quality--cost curves from one training run, avoids severe low-budget collapse in the main scaling experiment, achieves strong LRA and audio results, and substantially reduces multi-budget training cost in D4RL while remaining competitive with independently trained baselines. Future work includes larger-scale training, hybrid elastic architectures, and further efficiency optimization of the spectral branch; limitations and broader impacts are discussed in Appendix~\ref{app:limitations_impacts}.

\begin{ack}
This work was conducted at the Multi-agent Intelligence, Control, and Optimization (MICO) Lab at George Mason University. The MICO Lab is supported in part by the Army Research Office under award W911NF2220242 and the National Science Foundation under awards CMMI-DCSD-2540134 and ECCS-EPCN-2332210.
\end{ack}

\bibliographystyle{unsrtnat}
\bibliography{reference}






\appendix



\newpage

\section{Limitations and broader Impacts}
\label{app:limitations_impacts}

\softparagraph{Limitations.}
ES-SSM is evaluated up to the around-$1.3$B-parameter, $30$B-byte-token SlimPajama setting and across several long-sequence benchmarks.
Extending ES-SSM to larger multi-billion-parameter models and trillion-token training regimes, including standard large language model and hybrid architectures, is a natural next step for future work. The current experiments also consider a fixed set of runtime budgets $\mathcal{K}$: finer-grained budget choices and online budget selection have the potential to improve deployment flexibility. Finally, compared with Transformer attention, which benefits from mature fused implementations and inference-system support, ES-SSM still has room for systems-level efficiency optimization, including FFT batching, memory layout, projection fusion, and backend support.


\softparagraph{Broader impacts.}
ES-SSM aims to reduce deployment cost by exporting multiple compact models from one full-capacity training run, which may lower the compute and energy needed to support heterogeneous hardware budgets. This can make long-sequence modeling more accessible on resource-constrained platforms, but the same efficiency gains can also lower the barrier to deploying sequence models in harmful or poorly governed applications. ES-SSM does not directly address dataset bias, privacy leakage, security risks, or misuse; deployments should inherit the safeguards, data-governance practices, and monitoring appropriate to the downstream application.

\section{Additional derivations and proofs}
\label{app:derivations_proofs}
\subsection{Derivation of the Hankel spectral parameterization}
\label{app:spectral_derivation}

This appendix derives the base spectral parameterization used by ES-SSM:
a stable SSM kernel is expanded over fixed Hankel spectral channels as
$G(\tau)\approx\sum_{k=1}^{K}M_k\phi_k(\tau)$, yielding the corresponding
input-output map $\hat y(t)=\sum_{k=1}^{K}M_k(\Phi_k*u)(t)$. The main text presents the key idea:
a stable SSM kernel can be decomposed into scalar geometric traces, and these
traces can be approximated by a small number of fixed Hankel spectral channels.
Here we give the detailed construction and notation.

\subsection{From SSM eigenmodes to geometric traces}
\label{app:from_eigenmodes_to_traces}

Consider the causal convolutional kernel induced by a linear state-space model,
\[
    G(\tau)=CA^\tau B,
    \qquad \tau=0,\ldots,L-1,
\]
where
\[
    A\in\mathbb{R}^{N\times N},\qquad
    B\in\mathbb{R}^{N\times d_{\mathrm{in}}},\qquad
    C\in\mathbb{R}^{d_{\mathrm{out}}\times N}.
\]
Here $N$ is the hidden state dimension and $L$ is the finite convolution
horizon.

For clarity, we follow the standard stable symmetric setting used in spectral
filtering analyses~\citep{hazan2017learning,agarwal2023spectral}. Assume that
$A$ is real symmetric and has eigenvalues in $[0,1)$. Then
\[
    A=U\Lambda U^\top,
    \qquad
    \Lambda=\mathrm{diag}(\lambda_1,\ldots,\lambda_N),
\]
where $U$ is orthogonal and $\lambda_\ell\in[0,1)$. Substituting this
decomposition into the kernel gives
\[
    G(\tau)
    =
    C U \Lambda^\tau U^\top B.
\]
Expanding over the eigenmodes yields
\begin{equation}
    G(\tau)
    =
    \sum_{\ell=1}^{N}
    \lambda_\ell^\tau
    (CU)_{:\ell}(U^\top B)_{\ell:}.
    \label{eq:app_G_diag_decomp}
\end{equation}

Define the input-output coefficient matrix associated with the $\ell$-th
eigenmode as
\begin{equation}
    W_\ell
    \triangleq
    (CU)_{:\ell}(U^\top B)_{\ell:}
    \in
    \mathbb{R}^{d_{\mathrm{out}}\times d_{\mathrm{in}}}.
    \label{eq:app_W_def}
\end{equation}
Then \cref{eq:app_G_diag_decomp} can be written compactly as
\begin{equation}
    G(\tau)
    =
    \sum_{\ell=1}^{N}
    \lambda_\ell^\tau W_\ell.
    \label{eq:app_G_W_form}
\end{equation}

This expression shows that the matrix-valued kernel is a sum of scalar temporal
traces multiplied by input-output coefficient matrices. For each eigenvalue
$\lambda_\ell$, the time-dependent part is the geometric sequence
\[
    [1,\lambda_\ell,\lambda_\ell^2,\ldots,\lambda_\ell^{L-1}].
\]
Thus approximating the SSM kernel over the horizon $L$ reduces to
approximating scalar geometric sequences.

For a generic $\lambda\in[0,1)$, define the geometric vector
\begin{equation}
    \nu(\lambda)
    \triangleq
    \big[
    1,\lambda,\lambda^2,\ldots,\lambda^{L-1}
    \big]^\top
    \in\mathbb{R}^{L}.
    \label{eq:app_nu_def}
\end{equation}
Its entries satisfy
\[
    [\nu(\lambda_\ell)]_{\tau+1}
    =
    \lambda_\ell^\tau,
    \qquad \tau=0,\ldots,L-1.
\]
Therefore, although $\nu(\lambda_\ell)$ does not appear explicitly in
\cref{eq:app_G_W_form}, the scalar trace
$\{\lambda_\ell^\tau\}_{\tau=0}^{L-1}$ is exactly the entrywise
representation of $\nu(\lambda_\ell)$.

\subsection{Construction of Hankel spectral channels}
\label{app:hankel_channel_construction}

Following the spectral filtering construction of
\citet{hazan2017learning}, we build a universal basis for stable geometric
traces using a reweighted version of $\nu(\lambda)$. Define
\begin{equation}
    \mu(\lambda)
    \triangleq
    (1-\lambda)\nu(\lambda)
    =
    (1-\lambda)
    \big[
    1,\lambda,\lambda^2,\ldots,\lambda^{L-1}
    \big]^\top,
    \qquad \lambda\in[0,1].
    \label{eq:app_mu_def}
\end{equation}
The endpoint $\lambda=1$ is included only for the reweighted vector
$\mu(\lambda)$, for which $\mu(1)=0$. The factor $1-\lambda$ keeps the
family $\{\mu(\lambda)\}$ uniformly bounded near the stability boundary.

The Hankel matrix used to construct the spectral channels is
\begin{equation}
    Z
    \triangleq
    \int_{0}^{1}
    \mu(\rho)\mu(\rho)^\top
    d\rho
    \in\mathbb{R}^{L\times L},
    \label{eq:app_hankel_Z_def}
\end{equation}
where $\rho$ is a dummy integration variable. Equivalently, its entries are
\[
    Z_{ij}
    =
    \int_0^1
    (1-\rho)^2 \rho^{i+j-2}
    d\rho,
    \qquad i,j=1,\ldots,L.
\]
Thus $Z$ is a positive semidefinite Hankel matrix.

Let
\[
    \sigma_1\ge \sigma_2\ge\cdots\ge\sigma_L\ge 0
\]
denote the eigenvalues of $Z$, and let
\[
    \phi_1,\phi_2,\ldots,\phi_L\in\mathbb{R}^{L}
\]
denote the corresponding normalized eigenvectors. We refer to
$\phi_k$ as the $k$-th Hankel spectral channel. These channels depend only
on the sequence length $L$, not on the hidden dimension $N$ or on a
particular realization of $A,B,C$.

Using the first $K$ channels gives the top-$K$ projection
\begin{equation}
    \tilde{\mu}_K(\lambda)
    \triangleq
    \sum_{k=1}^{K}
    \langle \mu(\lambda),\phi_k\rangle \phi_k
    \in\mathbb{R}^{L}.
    \label{eq:app_mu_projection}
\end{equation}
Lemma 4.1 and the spectral decay bound in Appendix E of
\citet{hazan2017learning} imply that, for $L\ge 10$, $1\le K\le L$,
and a universal constant $c_{\mathrm H}>0$,
\begin{equation}
    \bigl\|
    \mu(\lambda)-\tilde{\mu}_K(\lambda)
    \bigr\|_2
    \le
    c_{\mathrm H}(\log L)^{1/4}
    \exp\!\left(
    -\frac{\pi^2 K}{16\log L}
    \right),
    \qquad \lambda\in[0,1].
    \label{eq:app_mu_proj_error}
\end{equation}
The cited reconstruction result bounds the squared norm by
$\bigl(6\sum_{j>K}\sigma_j\bigr)^{1/2}$; taking its square root yields
the norm bound above. Thus the leading Hankel channels provide an
exponentially improving approximation as $K$ increases.

\subsection{From reweighted traces to geometric traces}
\label{app:mu_to_nu_transfer}

The bound in \cref{eq:app_mu_proj_error} is stated for the reweighted vector
$\mu(\lambda)=(1-\lambda)\nu(\lambda)$. To relate it to the original
geometric vector $\nu(\lambda)$, define
\begin{equation}
    \tilde{\nu}_K(\lambda)
    \triangleq
    \sum_{k=1}^{K}
    \langle \nu(\lambda),\phi_k\rangle \phi_k,
    \qquad \lambda\in[0,1).
    \label{eq:app_nu_projection}
\end{equation}
Because projection onto the span of $\{\phi_1,\ldots,\phi_K\}$ is linear and
\[
    \mu(\lambda)=(1-\lambda)\nu(\lambda),
\]
we have
\begin{equation}
    \tilde{\mu}_K(\lambda)
    =
    (1-\lambda)\tilde{\nu}_K(\lambda).
    \label{eq:app_mu_nu_projection_relation}
\end{equation}
Therefore,
\begin{equation}
    \bigl\|
    \nu(\lambda)-\tilde{\nu}_K(\lambda)
    \bigr\|_2
    =
    \frac{1}{1-\lambda}
    \bigl\|
    \mu(\lambda)-\tilde{\mu}_K(\lambda)
    \bigr\|_2.
    \label{eq:app_nu_transfer}
\end{equation}

Consequently, the approximation guarantee for $\mu(\lambda)$ transfers to
$\nu(\lambda)$ on any interval $[0,1-\delta]$ bounded away from the
stability boundary. In particular, for $\lambda\le 1-\delta$,
\[
\|\nu(\lambda)-\tilde\nu_K(\lambda)\|_2
\le
\delta^{-1}c_{\mathrm H}(\log L)^{1/4}
\exp\!\left(-\frac{\pi^2 K}{16\log L}\right).
\]
The factor $1/(1-\lambda)$ reflects the fact that
unweighted geometric traces become harder to approximate uniformly as
$\lambda\to 1$.

\subsection{Substitution into the SSM kernel}
\label{app:substitution_into_kernel}

We now substitute the spectral approximation of each scalar geometric trace
into the SSM kernel. From \cref{eq:app_nu_projection}, for each eigenvalue
$\lambda_\ell$,
\[
    \nu(\lambda_\ell)
    \approx
    \sum_{k=1}^{K}
    \langle \nu(\lambda_\ell),\phi_k\rangle \phi_k.
\]
Taking the $(\tau+1)$-th entry gives
\begin{equation}
    \lambda_\ell^\tau
    =
    [\nu(\lambda_\ell)]_{\tau+1}
    \approx
    \sum_{k=1}^{K}
    \langle \nu(\lambda_\ell),\phi_k\rangle
    [\phi_k]_{\tau+1}.
    \label{eq:app_scalar_entry_expansion}
\end{equation}
For notational convenience, define
\[
    \phi_k(\tau)
    \triangleq
    [\phi_k]_{\tau+1},
    \qquad \tau=0,\ldots,L-1.
\]
Then
\begin{equation}
    \lambda_\ell^\tau
    \approx
    \sum_{k=1}^{K}
    \langle \nu(\lambda_\ell),\phi_k\rangle
    \phi_k(\tau).
    \label{eq:app_scalar_mode_expand}
\end{equation}

Substituting \cref{eq:app_scalar_mode_expand} into
\cref{eq:app_G_W_form} yields
\[
\begin{aligned}
    G(\tau)
    &=
    \sum_{\ell=1}^{N}
    \lambda_\ell^\tau W_\ell
    \approx
    \sum_{\ell=1}^{N}
    \sum_{k=1}^{K}
    \langle \nu(\lambda_\ell),\phi_k\rangle
    \phi_k(\tau) W_\ell.
\end{aligned}
\]
Exchanging the order of summation gives
\begin{equation}
\begin{aligned}
    G(\tau)
    &\approx
    \sum_{k=1}^{K}
    \left(
    \sum_{\ell=1}^{N}
    \langle \nu(\lambda_\ell),\phi_k\rangle W_\ell
    \right)
    \phi_k(\tau).
\end{aligned}
\label{eq:app_G_expand_intermediate}
\end{equation}
Define
\begin{equation}
    M_k
    \triangleq
    \sum_{\ell=1}^{N}
    \langle \nu(\lambda_\ell),\phi_k\rangle W_\ell
    \in
    \mathbb{R}^{d_{\mathrm{out}}\times d_{\mathrm{in}}}.
    \label{eq:app_Mk_def}
\end{equation}
Then the SSM kernel admits the spectral approximation
\begin{equation}
    G(\tau)
    \approx
    \sum_{k=1}^{K}
    M_k\phi_k(\tau),
    \qquad \tau=0,\ldots,L-1.
    \label{eq:app_G_expand}
\end{equation}

Finally, viewing
\[
    \Phi_k=\{\phi_k(\tau)\}_{\tau=0}^{L-1}
\]
as a causal convolution filter, the corresponding input-output map is
\begin{equation}
    \hat y(t)
    =
    \sum_{k=1}^{K}
    M_k(\Phi_k*u)(t),
    \label{eq:app_base_spectral_ssm}
\end{equation}
where $*$ denotes causal convolution.

In neural sequence modeling, we do not explicitly identify
$A,B,C,U$, or the eigenvalues $\{\lambda_\ell\}_{\ell=1}^{N}$.
Instead, the Hankel channels $\{\Phi_k\}_{k=1}^{K}$ are fixed in advance,
and the matrices $\{M_k\}_{k=1}^{K}$ are learned directly from data.
Thus the hidden dimension $N$ of the classical SSM realization is absorbed
into the learned channel projections, while the spectral budget $K$ becomes
the controllable inference budget.

\subsection{Prefix consistency of exported compact models}
\label{app:prefix_consistency}

We formalize the mechanical consistency property used by direct prefix
truncation. The statement is per spectral sublayer and conditional on the same
sublayer input sequence $u_{1:L}$; across a full network, different runtime
budgets may of course produce different intermediate representations after the
first budgeted layer.

\begin{remark}[Prefix consistency]
\label{rem:prefix_consistency}
Consider a full-capacity ES-SSM spectral sublayer with maximum budget
$\overline K$. Let the budget-$K$ exported compact sublayer be obtained by
retaining $\{\Phi_k,M_k\}_{k=1}^{K}$, the first $K$ rows of
$\mathbf{W}_{g,2}$, and the first $K$ entries of $\mathbf{b}_{g,2}$, while
keeping the shared gate parameters $\mathbf{W}_{g,1},\mathbf{b}_{g,1}$
unchanged. For any retained channel $k\le K$ and any token position $t$, the
exported sublayer computes the same gate logit $s_k(t)$, the same spectral
feature $(\Phi_k*u)(t)$, and uses the same projection matrix $M_k$ as the
full-capacity sublayer. Consequently, the budget-$K$ sublayer output is exactly
\[
[\mathcal{S}_K(u)](t)
=
\sum_{k=1}^{K}
\frac{1}{\sqrt K}\operatorname{sigmoid}(s_k(t))\,M_k(\Phi_k*u)(t),
\]
with no dependence on any removed channel $j>K$.
\end{remark}

\noindent\emph{Justification.}
Write the shared hidden gate activation as
\[
h_g(t)=\mathrm{GELU}(\mathbf{W}_{g,1}u(t)+\mathbf{b}_{g,1}).
\]
In the full-capacity sublayer, the $k$-th gate logit is
\[
s_k(t)=(\mathbf{W}_{g,2})_{k:}h_g(t)+(\mathbf{b}_{g,2})_k.
\]
For an exported budget-$K$ sublayer with $k\le K$, row $k$ of
$\mathbf{W}_{g,2}$ and entry $k$ of $\mathbf{b}_{g,2}$ are retained, and
$h_g(t)$ is computed from the unchanged shared parameters. Hence the value of
$s_k(t)$ is unchanged by removing rows $j>K$. The spectral filter $\Phi_k$ and
projection matrix $M_k$ are also retained unchanged, so $(\Phi_k*u)(t)$ and
$M_k(\Phi_k*u)(t)$ are identical for the retained channel. Substituting these
unchanged retained-channel quantities into the definition of
\cref{eq:adaptive_spectral_K} gives the displayed expression. Since all terms
with $j>K$ are absent from the exported sum and no retained logit or projection
depends on those removed channel-indexed tensors, the exported computation has
no dependence on removed channels.

\subsection{Gradient allocation under budget dropout}
\label{app:budget_dropout_gradients}

We consider the default step-level sampling schedule. The training schedule also has a simple consequence for which channel-specific
parameters enter the forward pass and can receive task gradients. Let
\[
\Theta_k^{\mathrm{ch}}
\;=\;
\{M_k,(\mathbf{W}_{g,2})_{k:},(\mathbf{b}_{g,2})_k\}
\]
denote the trainable parameters associated specifically with spectral channel
$k$ in one ES-SSM sublayer. The fixed Hankel filter $\Phi_k$ is not included
because it is a non-learnable buffer.

\begin{remark}[Budget-dropout update allocation]
\label{rem:budget_dropout_gradients}
At a training step with sampled budget $K_{\mathrm{train}}$, the parameters
$\Theta_k^{\mathrm{ch}}$ enter the computational graph only if
$k\le K_{\mathrm{train}}$. Therefore, channel-specific parameters for
lower-index channels have gradient paths at least as often as those for
higher-index channels. After warmup, if the full-budget anchor cadence
$n_{\mathrm{full}}$ is finite, every channel-specific parameter group
$\Theta_k^{\mathrm{ch}}$ is included in the forward pass on at least one out of
every $n_{\mathrm{full}}$ consecutive optimization steps.
\end{remark}

\noindent\emph{Justification.}
For a sampled budget $K_{\mathrm{train}}$, each spectral sublayer evaluates
\[
[\mathcal{S}_{K_{\mathrm{train}}}(u)](t)
=
\sum_{j=1}^{K_{\mathrm{train}}}
g_j^{(K_{\mathrm{train}})}(t)\,M_j(\Phi_j*u)(t).
\]
Only terms with $j\le K_{\mathrm{train}}$ appear in the forward computation.
Thus for any $k>K_{\mathrm{train}}$, the tensors in
$\Theta_k^{\mathrm{ch}}$ are not used to compute the loss at that step, so
their task gradient from that step is zero. Conversely, when
$k\le K_{\mathrm{train}}$, the corresponding projection, gate-output row, and
gate-output bias entry participate in the retained-channel computation and have
the usual backpropagation path.

Let $p_k=\Pr(K_{\mathrm{train}}\ge k)$ denote the post-warmup probability that
channel $k$ is included. Since the events $\{K_{\mathrm{train}}\ge k\}$ are
nested, $p_i\ge p_j$ whenever $i<j$, so lower-index channel-specific
parameters have gradient paths at least as often in expectation. If $q$ denotes
the distribution over budgets on non-anchor steps, then for finite
$n_{\mathrm{full}}$ the long-run inclusion probability is
\[
p_k
=
\frac{1}{n_{\mathrm{full}}}
+
\left(1-\frac{1}{n_{\mathrm{full}}}\right)
\sum_{\kappa\in\mathcal{K}:\,\kappa\ge k} q(\kappa).
\]
In particular, the anchor rule sets
$K_{\mathrm{train}}=\overline K$ whenever
$j\bmod n_{\mathrm{full}}=0$. Hence every block of
$n_{\mathrm{full}}$ consecutive post-warmup optimization steps contains a
full-budget step, which includes every channel $k\le\overline K$. This gives
each channel-specific parameter group an inclusion-frequency lower bound of
$1/n_{\mathrm{full}}$ from the anchor updates alone.

\section{Efficient implementation}
\label{app:efficient_impl}
This section describes the implementation choices used for the ES-SSM results reported in the main text.

\subsection{Exported inference and budget-dropout training pseudocode}
\label{app:method_algorithms}

\begin{algorithm}[t]
\caption{ES-SSM inference}
\label{alg:inference}
\SetAlgoLined
\DontPrintSemicolon
\KwIn{input $\mathbf{u}$; trained full-capacity parameters $\theta$; runtime budget $K\le\overline K$}
\KwOut{output $\hat{\mathbf{y}}^{(K)}$}

\textit{Export} $\theta^{(K)}$ by direct prefix truncation in every ES-SSM spectral sublayer:
retain $\{\Phi_k,M_k\}_{k=1}^{K}$, the first $K$ rows of $\mathbf{W}_{g,2}$, and the first
$K$ entries of $\mathbf{b}_{g,2}$; remove channels $k>K$\;
\textit{Keep} all shared parameters unchanged and disable DropPath for deterministic inference\;
$\mathbf{r}\leftarrow \mathrm{Embed}(\mathbf{u})$\;
\For{each ES-SSM block $\ell$ in $\theta^{(K)}$}{
  $\mathbf{v}\leftarrow \mathrm{LN}^{(\ell)}_1(\mathbf{r})$\;
  $\mathbf{o}^{(\ell)}\leftarrow \mathcal{S}^{(\ell)}_K(\mathbf{v})$\;
  \tcp*{spectral sublayer}
  $\mathbf{r}_{\mathrm{mid}}\leftarrow \mathbf{r}+\mathbf{o}^{(\ell)}$\;
  $\mathbf{r}\leftarrow \mathbf{r}_{\mathrm{mid}}+
  \mathrm{FFN}^{(\ell)}(\mathrm{LN}^{(\ell)}_2(\mathbf{r}_{\mathrm{mid}}))$\;
}
$\hat{\mathbf{y}}^{(K)}\leftarrow \mathrm{OutputHead}(\mathbf{r})$\;
\Return $\hat{\mathbf{y}}^{(K)}$\;
\end{algorithm}

\begin{algorithm}[t]
\caption{ES-SSM training}
\label{alg:training}
\SetAlgoLined
\DontPrintSemicolon
\KwIn{data $\mathcal{D}$; runtime-budget set $\mathcal{K}$; $\overline K$; $E_{\mathrm{warmup}}$; $n_{\mathrm{full}}$}
\KwOut{trained parameters $\theta$}

$K_{\min}\leftarrow \min\mathcal{K}$\;
\For{each minibatch $(\mathbf{u},\mathbf{y})\sim\mathcal{D}$ at epoch $e$ and step $j$}{
  \eIf{$e < E_{\mathrm{warmup}} \;\lor\;
  (n_{\mathrm{full}}<\infty \wedge j\bmod n_{\mathrm{full}}=0)$}{
    $K_{\mathrm{train}}\leftarrow \overline K$\tcp*{warmup / anchor}
  }{
    draw $\xi\sim\mathrm{Unif}[0,1]$\;
    $K_{\mathrm{target}}\leftarrow K_{\min}(\overline K/K_{\min})^{\xi}$\;
    $K_{\mathrm{train}}\leftarrow
    \arg\min_{k\in\mathcal{K}}|k-K_{\mathrm{target}}|$\;
  }
  $\hat{\mathbf{y}}^{(K_{\mathrm{train}})}
  \leftarrow F_{\theta}(\mathbf{u};K_{\mathrm{train}})$\;
  $\mathcal{L}\leftarrow
  \mathcal{L}_{\mathrm{task}}(\hat{\mathbf{y}}^{(K_{\mathrm{train}})},\mathbf{y})$\;
  update $\theta$ with AdamW, $\nabla_{\theta}\mathcal{L}$, and global
  $\ell_2$-norm gradient clipping\;
}
\Return $\theta$\;
\end{algorithm}

\subsection{FFT-based spectral convolution}
\label{app:fft}

The spectral features $\{(\Phi_k*u)(t)\}_{k=1}^{\overline K}$ are computed jointly by a single length-$n$ real FFT with $n=2L$, which is the standard zero-padded length required for a \emph{causal} linear convolution on sequences of length $L$.\footnote{Using $n=L$ would alias the tail of the convolution and is incorrect for causal filtering.}

Throughout this appendix we use the main-text setting
$d_{\mathrm{in}}\!=\!d_{\mathrm{out}}\!=\!d$. Let
$\mathcal{U}\in\mathbb{R}^{B_{\mathrm{sz}}\times d\times L}$ be the layer input with the time
axis last (batch size $B_{\mathrm{sz}}$, feature dimension $d$), and let
$\widehat{\mathcal{U}} = \mathrm{rFFT}(\mathcal{U},\,n=2L)\in\mathbb{C}^{B_{\mathrm{sz}}\times d\times(n/2+1)}$.
The filter transforms
$\{\widehat\Phi_k\}_{k=1}^{\overline K}
\in\mathbb{C}^{\overline K\times(n/2+1)}$ are precomputed once at
model construction (Appendix~\ref{app:hankel_filter_bank}) and stored as a
non-learnable buffer. Writing $c$ for the feature-channel index to
avoid collision with the dimension symbol $d$, the sublayer evaluates
\begin{equation}
\mathcal{F}_{b,t,k,c}
\;=\;
\mathrm{irFFT}\!\big(\widehat{\mathcal{U}}_{b,c}\,\odot\,\widehat\Phi_{k},\;n=2L\big)_{t},
\qquad t=1,\ldots,L,\;c=1,\ldots,d,
\label{eq:app_fft_output}
\end{equation}
so that $\mathcal{F}\in\mathbb{R}^{B_{\mathrm{sz}}\times L\times\overline K\times d}$, with the inverse FFT output truncated to its first $L$ samples to recover the causal convolution tail. All $\overline K$ channels share the same $\widehat{\mathcal{U}}$, so the full spectral feature bank is produced by one forward rFFT, one broadcast complex multiply of shapes $(B_{\mathrm{sz}},d,1,n/2+1)\times(1,1,\overline K,n/2+1)$, and one inverse rFFT followed by a transpose to the layout above.

For an exported compact model at runtime budget $K\le\overline K$, the broadcast multiply is built from the first $K$ filter transforms $\{\widehat\Phi_k\}_{k=1}^{K}$. Every downstream tensor ($\mathcal{F}$, the gate $g^{(K)}$, and the projection stack $\{M_k\}$) is likewise sliced along the channel axis. No recomputation or re-fitting is required.

\subsection{Precision policy}
\label{app:precision}

\softparagraph{Design.}
The bulk of the network (embeddings, attention/SSM matmuls, FFN,
LayerNorms) runs in \texttt{bf16} under PyTorch AMP. The spectral
branch in \cref{eq:app_fft_output}, however, runs in
\texttt{float32} with autocast explicitly disabled. The Hankel
eigenvalues span several orders of magnitude, so low-precision spectral
arithmetic can bias the smaller-magnitude channels. Keeping the FFT,
complex multiply, and inverse FFT in \texttt{fp32} removes this effect
at negligible cost, while \texttt{bf16} remains the default for the
rest of the network because it is stable without loss scaling.

\begin{figure}[t]
\centering
\includegraphics[width=0.85\textwidth]{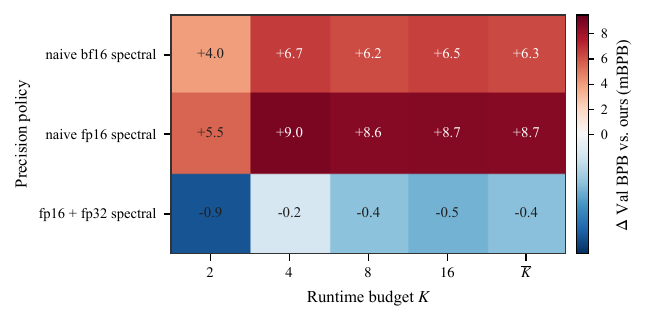}
\caption{$\Delta$ Val BPB relative to our policy (row D in
\cref{tab:precision_ablation}) at runtime budget
$K\in\{2,4,8,16,\overline K\}$. Positive entries (red) are worse than
ours; negative (blue) are better. Both naive low-precision rows (B, C)
are uniformly positive ($10/10$ cells), and the fp16-naive row C is
darker than the bf16-naive row B at every $K$. The forced-\texttt{fp32}
spectral row (E) is statistically indistinguishable from D
($|\Delta_K|<0.001$\,BPB at every $K$).}
\label{fig:precision_delta_heatmap}
\end{figure}

\begin{figure}[t]
\centering
\includegraphics[width=0.95\textwidth]{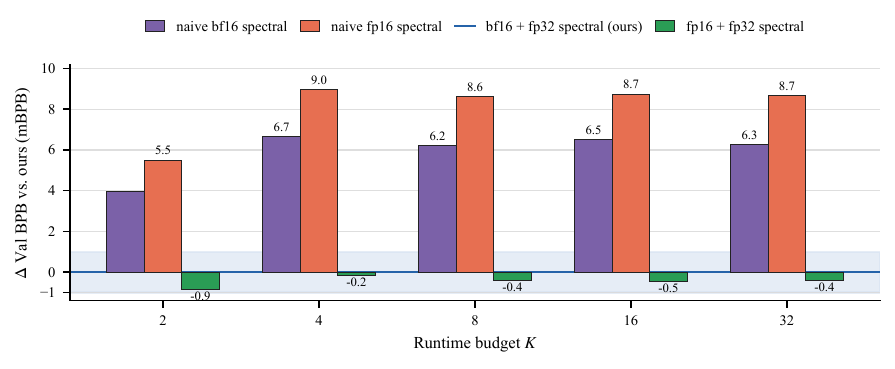}
\caption{Per-$K$ validation-BPB difference relative to our precision policy
(D) after $1.5$\,B SlimPajama tokens. Positive values are worse than our
policy. The naive low-precision spectral configurations are worse at every
runtime budget, while the fp16 main-network run with an fp32 spectral branch
stays within the $\pm 1$ mBPB band.}
\label{fig:precision_per_k_bars}
\end{figure}

\begin{figure}[t]
\centering
\includegraphics[width=0.95\textwidth]{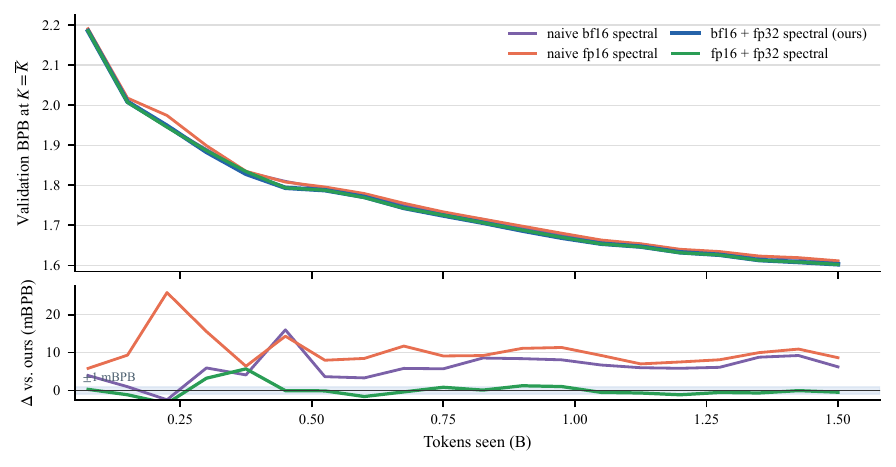}
\caption{Training dynamics for the precision-policy ablation. Top:
validation BPB at $K=\overline K=32$ over training. Bottom:
validation-BPB difference relative to our policy (D), in mBPB. The two
fp32-spectral configurations track each other within a small band, while
the naive low-precision spectral configurations keep a positive offset
after early training.}
\label{fig:precision_train_curves}
\end{figure}

\softparagraph{Empirical validation.}
We validate the policy with a controlled four-config ablation at
$160$\,M parameters on SlimPajama
($d{=}512$, $n_{\mathrm{layers}}{=}15$, $\overline K{=}32$, $L{=}2048$,
$1.5$\,B byte-tokens).
The four configurations differ only in (i)~the AMP main-network dtype
and (ii)~whether the spectral branch is forced to \texttt{fp32}; all
other hyperparameters and the random seed are identical.
\Cref{tab:precision_ablation} reports validation BPB at three runtime
budgets and the gap to our policy ($\Delta$) at $K{=}\overline K$.
\Cref{fig:precision_delta_heatmap,fig:precision_per_k_bars,fig:precision_train_curves}
visualise the same data three ways.

\begin{table}[t]
\centering
\small
\begin{tabular}{lcccccc}
\toprule
Config & main & spec.\ & Val BPB$_{K=2}$ & Val BPB$_{K=8}$ & Val BPB$_{K=\overline K}$ & $\Delta_{K=\overline K}$ \\
\midrule
B. naive bf16          & \texttt{bf16} & low           & 1.7826 & 1.6064 & 1.6089 & $+0.0063$ \\
C. naive fp16+scaler   & \texttt{fp16} & low           & 1.7841 & 1.6088 & 1.6114 & $+0.0087$ \\
D. \texttt{bf16}+\texttt{fp32} spectral (\textbf{ours}) & \texttt{bf16} & \texttt{fp32} & 1.7786 & 1.6002 & 1.6027 & \,---     \\
E. \texttt{fp16}+\texttt{fp32} spectral                & \texttt{fp16} & \texttt{fp32} & 1.7778 & 1.5997 & 1.6023 & $-0.0004$ \\
\bottomrule
\end{tabular}
\caption{Precision-policy ablation at $160$\,M parameters on SlimPajama,
$1.5$\,B byte-tokens, $\overline K{=}32$. All four configurations share
architecture, optimizer, schedule, data, and random seed; only the AMP
main-network dtype and whether the spectral branch is forced to
\texttt{fp32} differ. ``low'' in the spec.\ column means the spectral
multiply runs in the AMP main dtype (i.e.~the FFT itself is auto-cast to
\texttt{fp32} by PyTorch's AMP fallback for FFT, but the complex multiply
runs in \texttt{bf16}/\texttt{fp16}). $\Delta_{K=\overline K}$ is val BPB
minus our policy (D) at $K{=}\overline K{=}32$.}
\label{tab:precision_ablation}
\end{table}

The ablation supports our precision policy as a practical
accuracy--efficiency choice in this regime. Both naive low-precision
spectral configurations are worse at every runtime budget: the
full-capacity penalty is $+0.006$\,BPB for bf16 spectral arithmetic and
$+0.009$\,BPB for fp16 spectral arithmetic. By contrast, once the
spectral branch is forced to \texttt{fp32}, the bf16-main and fp16-main
runs are indistinguishable within $0.001$\,BPB at every $K$. The training
curves show the same pattern throughout optimization rather than only at
the final checkpoint. We therefore use \texttt{bf16} for the main
network and \texttt{fp32} for the spectral branch.
\softparagraph{Cost.}
Using \texttt{fp32} in the spectral branch does not add measurable
wall-time overhead in this experiment: the two \texttt{fp32}-spectral
runs finish within measurement noise of each other. The naive
low-precision spectral emulation is actually slower, because complex
half-precision multiplication must be manually decomposed into real and
imaginary components under PyTorch AMP.
\softparagraph{Limitations.}
This is a single-seed ablation at $1.5$\,B tokens. The observed
low-precision spectral penalties are larger than nearby validation noise,
but we treat the experiment as evidence for the precision policy rather
than as a full scaling law for numerical precision.

\subsection{Gate-before-projection ordering}
\label{app:gate_before_proj}

Given the spectral feature tensor
$\mathcal{F}\in\mathbb{R}^{B_{\mathrm{sz}}\times L\times\overline K\times d}$ from
\cref{eq:app_fft_output} and the gate coefficients
$g^{(K)}\in\mathbb{R}^{B_{\mathrm{sz}}\times L\times K}$, two algebraically
equivalent orderings are possible for
\cref{eq:adaptive_spectral_K}:
\begin{align*}
\text{(a) gate-first:}\quad
&\big[g^{(K)}_{b,t,k}\,\mathcal{F}_{b,t,k,\cdot}\big]\,M_k^{\!\top},\\
\text{(b) project-first:}\quad
&g^{(K)}_{b,t,k}\,\big[\mathcal{F}_{b,t,k,\cdot}\,M_k^{\!\top}\big].
\end{align*}
We use ordering (a) and combine projection with channel summation in a
single contraction, avoiding an explicit
$(B_{\mathrm{sz}},L,K,d)$ per-channel projected-output tensor.
For $d_{\mathrm{in}}=d_{\mathrm{out}}=d$, both orderings require
$B_{\mathrm{sz}}LKd$ scalar gate multiplications. Concretely, the kernel
implementing ordering~(a) is a single \texttt{einsum} of the form
\texttt{"blkd, kde $\to$ ble"} applied to the gated features and the
stacked projection tensor
$\widetilde M\in\mathbb{R}^{\overline K\times d\times d}$ defined by
$\widetilde M[k,c,e]=(M_k)_{e,c}$ for input and output feature indices
$c,e\in\{1,\ldots,d\}$ (i.e.\ $M_k^{\!\top}$ stacked along
the channel axis), prefix-sliced to $\widetilde M[:K]$ at runtime.

\subsection{Parameter initialization and shared modules}
\label{app:init}

The channel projections $\{M_k\}_{k=1}^{\overline K}$ are initialized elementwise i.i.d.\ from $\mathcal{N}(0,\,10^{-6})$, i.e.\ with standard deviation $10^{-3}$. At step $0$ the spectral branch output therefore has near-zero magnitude; combined with the near-closed gates described below, this means the network begins training as a pure residual plus FFN path, and the spectral branch must be earned during optimization. The gating MLP uses default PyTorch \texttt{nn.Linear} initialization for both layers, with one exception: the output-layer bias $\mathbf{b}_{g,2}$ is overridden to the constant $-2$, giving $\operatorname{sigmoid}(s_k(t))\!\approx\!0.12$ for every $k,t$ at initialization. The token embedding is drawn from $\mathcal{N}(0,\,0.02^{2})$ with padding index $0$ and is tied to the output head (standard weight tying). DropPath follows a layerwise linear schedule from $0$ at the first layer to the nominal maximum (e.g.\ $0.1$) at the top layer, and dropout with rate $0.1$ is applied to the embedding output and within the FFN.

\subsection{Hankel filter bank construction}
\label{app:hankel_filter_bank}

The filter bank $\{\Phi_k\}_{k=1}^{\overline K}$ is the top-$\overline K$
eigensystem of the Hankel matrix $Z$ defined in
\cref{eq:app_hankel_Z_def,eq:app_mu_def}. The entries of $Z$ admit the closed
form
\begin{equation}
Z_{i,j}
=
\int_{0}^{1}(\rho-1)^{2}\rho^{i+j-2}\,d\rho
=
\frac{2}{(i+j)^{3}-(i+j)},
\qquad i,j=1,\ldots,L,
\label{eq:app_Z_entries}
\end{equation}
so $Z$ is a Hankel matrix generated by the scalar sequence $h_{q}=2/(q^{3}-q)$, $q=2,\ldots,2L$, where $\rho$ is the integration variable. We compute the leading $\overline K$ eigenpairs $(\sigma_k,\phi_k)$ once per sequence length $L$ and cache them.

We never materialize $Z$ as a full $L\times L$ matrix. Instead, any Hankel matrix--vector product $Zv$ is computed in $\mathcal{O}(L\log L)$ by embedding $Z$ into a circulant of size $2L$: one forward rFFT on the zero-padded $v$, one elementwise complex multiply with the precomputed generating-vector transform $\widehat h=\mathrm{rFFT}(\tilde h)$, where $\tilde h$ is the circulant embedding of the generating sequence $\{h_q\}$, and one inverse rFFT to read off the relevant entries of the result. With oversampling margin $r_{\mathrm{os}}=16$, a Rayleigh--Ritz iteration applies $Z$ to the $\overline K+r_{\mathrm{os}}$ columns of $Q$ as a single batched matvec, so each iteration requires two batched FFTs in total.

We extract $\{(\sigma_k,\phi_k)\}_{k\le\overline K}$ via subspace
iteration with periodic Rayleigh--Ritz restarts:
(i)~initialize $Q\in\mathbb{R}^{L\times(\overline K+r_{\mathrm{os}})}$ with Gaussian
entries and orthonormalize via QR;
(ii)~repeat for $T_{\max}=60$ iterations:
    $Q\leftarrow\mathrm{QR}(ZQ)$, and every $\lfloor T_{\max}/4\rfloor$
    iterations perform a Rayleigh--Ritz step
    ($R\!=\!Q^\top Z Q$; eigendecompose $R$; sort descending;
    rotate $Q\leftarrow QV_R$, where $V_R$ contains the Ritz vectors) and check the per-eigenpair residual
    $\mathrm{res}_k=\|ZQ_{:,k}-\sigma_k Q_{:,k}\|_2/|\sigma_k|$;
(iii)~stop early when $\max_{k\le\overline K}\mathrm{res}_k<10^{-6}$.
The resulting top-$\overline K$ eigenvectors form
$\{\phi_k\}_{k=1}^{\overline K}$, and their rFFTs (padded to length
$2L$) form the filter-transform buffer used in
Appendix~\ref{app:fft}.

Because $\overline K$ and $L$ are fixed for a given experiment, the filter bank is computed once on CPU and cached to disk; subsequent runs load the cached tensor. This precomputation takes under a minute for $L=2048,\overline K=32$ and is not counted in training wall-clock.

\subsection{Full training recipe}
\label{app:training_recipe}

\Cref{tab:training_recipe} lists the optimization hyperparameters used
in all main-text experiments unless stated otherwise.

\begin{table}[t]
\centering
\small
\setlength{\tabcolsep}{6pt}
\renewcommand{\arraystretch}{1.15}
\caption{Training recipe.}
\label{tab:training_recipe}
\begin{tabular}{ll}
\toprule
Hyperparameter           & Value \\
\midrule
Optimizer                & AdamW, $(\beta_1,\beta_2)\!=\!(0.9,0.95)$, $\varepsilon\!=\!10^{-8}$ \\
Weight decay             & $0.1$ on non-LayerNorm, non-embedding, non-bias parameters \\
Peak learning rate       & $3\!\times\!10^{-4}$ \\
Warmup                   & Linear, first $2\%$ of total optimization steps \\
Decay                    & Cosine to $10\%$ of peak at the final step \\
Gradient clipping        & Global $\ell_{2}$ norm clipped to $1.0$ \\
Effective batch size     & Scale-dependent; see \cref{sec:experiments} \\
Mixed precision          & See Appendix~\ref{app:precision} \\
Activation checkpointing & Enabled for large SlimPajama-scale ES-SSM runs; off for small ablations \\
Early stopping           & Val.\ BPB at $K\!=\!\overline K$; patience $8$ epochs \\
\bottomrule
\end{tabular}

\end{table}

\section{Budget dropout sampler}
\label{app:budget_sampler}

This appendix specifies the sampler referred to in the main text (\cref{subsec:budget_adaptive_training}, \cref{eq:budget_sampler}). It implements three design constraints: (i) early training should resemble full-capacity training so that every per-channel parameter is given a chance to initialize meaningfully; (ii) throughout training, the tail channels $k$ close to $\overline K$ must continue to receive gradient signal, or they will drift and the full-capacity model will degrade; and (iii) the mid-training distribution over $K_{\mathrm{train}}$ should cover $[K_{\min},\overline K]$ approximately uniformly on the log-scale so that every runtime budget in the set is exercised at comparable frequency.

\begin{algorithm}[t]
\caption{\textsc{SampleBudget}: budget dropout sampler used during training}
\label{alg:budget_sampler}
\SetAlgoLined
\DontPrintSemicolon
\KwIn{runtime-budget set $\mathcal{K}$ with $\overline K=\max\mathcal{K}$; epoch $e$; global optimization step $j$; warmup epochs $E_{\mathrm{warmup}}$; full-budget anchor cadence $n_{\mathrm{full}}$; random generator \textsc{rng}}
\KwOut{sampled training budget $K_{\mathrm{train}}\in\mathcal{K}$}

$K_{\min}\leftarrow \min\mathcal{K}$\;
\eIf{$e < E_{\mathrm{warmup}}$}{
  $K_{\mathrm{train}}\leftarrow \overline K$
  \tcp*{full-budget warmup}
}{
  \eIf{$n_{\mathrm{full}}<\infty$ and $j \bmod n_{\mathrm{full}}=0$}{
    $K_{\mathrm{train}}\leftarrow \overline K$
    \tcp*{full-budget anchor update}
  }{
    draw $\xi\sim\mathrm{Unif}[0,1]$ using \textsc{rng}\;
    $K_{\mathrm{target}}\leftarrow K_{\min}(\overline K/K_{\min})^{\xi}$\tcp*{log-scale target}
    $K_{\mathrm{train}}\leftarrow
    \arg\min_{k\in\mathcal{K}}|k-K_{\mathrm{target}}|$\tcp*{snap to nearest allowed budget}
  }
}
\Return $K_{\mathrm{train}}$\;
\end{algorithm}

\softparagraph{Default hyperparameters.} Unless a benchmark-specific setup states otherwise, we use $E_{\mathrm{warmup}}\!=\!8$ epochs, $n_{\mathrm{full}}\!=\!8$ steps, $K_{\min}\!=\!2$, $\overline K\!=\!32$, and $\mathcal{K}\!=\!\{2,3,4,5,6,8,12,16,24,32\}$. The RNG is seeded deterministically so that
two runs with the same seed and sampling protocol receive identical
$K_{\mathrm{train}}$ sequences.

Two choices in \cref{alg:budget_sampler} control gradient allocation
across spectral channels. The warmup phase keeps
$K_{\mathrm{train}}\!=\!\overline K$ while the small initialized
projections and near-closed gates settle into a stable regime. After
warmup, the full-budget anchor update guarantees every channel a minimum
update rate of $1/n_{\mathrm{full}}$, preventing tail channels from being
trained only through rare sampled full-budget steps.

\softparagraph{Empirical validation of the full-budget anchor cadence.}
The default $n_{\mathrm{full}}\!=\!8$ lies in the empirical sweet spot
of a more detailed cadence sweep. Let $p_{\max}$ be the probability that the
log-scale sampler selects $\overline K$ before the full-budget anchor
rule is applied; for our runtime-budget set, $p_{\max}\approx0.048$.
The per-step probability of training at $K\!=\!\overline K$ is
\begin{equation}
\label{eq:nfull_prob}
P(K_{\mathrm{train}}\!=\!\overline K)
\;=\;
\frac{1}{n_{\mathrm{full}}}
\;+\;
\Big(1-\frac{1}{n_{\mathrm{full}}}\Big)\!
\cdot p_{\max}
\;=\;
\frac{1}{n_{\mathrm{full}}}
\;+\;
\frac{n_{\mathrm{full}}-1}{n_{\mathrm{full}}}\,p_{\max}.
\end{equation}
For $n_{\mathrm{full}}\!=\!8$, this is approximately $0.167$, so the
maximum-budget path receives a full-capacity update roughly once every
six optimizer steps. Smaller cadences over-emphasize the full-budget path,
while larger cadences leave tail channels too dependent on the sampled
log-scale distribution.

\begin{figure}[t]
\centering
\includegraphics[width=\linewidth]{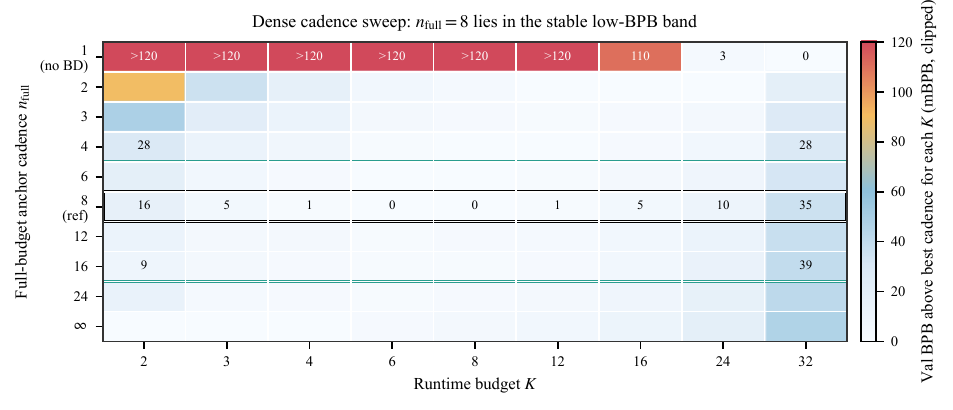}
\caption{Full-budget anchor cadence sweep.
Each cell shows validation BPB above the best cadence for that runtime budget,
in milli-BPB and clipped at $120$ for readability. The black row marks the
default $n_{\mathrm{full}}\!=\!8$, which forces
$K_{\mathrm{train}}\!=\!\overline K$ once every eight post-warmup optimizer
steps; the green box marks the empirically stable band
$n_{\mathrm{full}}\!\in\!\{4,6,8,12,16\}$. The endpoint
$n_{\mathrm{full}}\!=\!1$ degenerates to full-budget training and collapses
small-$K$ prefixes, while very large cadences undertrain the tail channels.
Exact means and standard deviations are reported in
\cref{tab:abl_full_every_validation}.}
\label{fig:full_budget_anchor_cadence}
\end{figure}

The sweep in \cref{fig:full_budget_anchor_cadence,tab:abl_full_every_validation}
tests $n_{\mathrm{full}}\!\in\!\{1,2,3,4,6,8,12,16,24,\infty\}$ with
three seeds per cadence. The coarser table in Appendix~\ref{app:abl_full_every}
is retained only to show that the endpoint $n_{\mathrm{full}}\!=\!1$
recovers the no-budget-dropout failure mode, while reasonable
non-endpoint cadences behave similarly.

\begin{table}[t]
\centering\small
\setlength{\tabcolsep}{4pt}
\renewcommand{\arraystretch}{1.05}
\caption{Effect of the full-budget anchor cadence
$n_{\mathrm{full}}$ on validation BPB across runtime budgets,
mean$\,\pm\,$std over $3$ random seeds using a $287.1$\,M-parameter
model trained for $80$\,M tokens per run
($\approx\!0.28$ tokens per parameter). Two competing effects
make the curve U-shaped: as $n_{\mathrm{full}}\!\downarrow$ the
$K\!=\!\overline K$ channel gets more gradient and $K\!=\!\overline K$
BPB falls monotonically; as $n_{\mathrm{full}}\!\downarrow$ the
small-$K$ operating points get less probability mass and small-$K$
BPB rises. The reference choice $n_{\mathrm{full}}\!=\!8$ sits in
a wide flat optimum spanning $n_{\mathrm{full}}\!\in\![4,16]$ at
every $K$, inside which the trade-off is empirically balanced —
$K\!=\!2$ val BPB lies in $[2.447,2.466]$ and $K\!=\!\overline K$
val BPB in $[2.215,2.226]$, all within two standard errors of
one another. The catastrophic collapse at $n_{\mathrm{full}}\!=\!1$
($K\!=\!2$ val BPB $4.245\!\pm\!0.190$) recovers the
no-budget-dropout failure mode of \cref{tab:abl_bd} as a degenerate
limit. The sweep does not identify a single best
$n_{\mathrm{full}}$ — neither $4$ nor $16$ is statistically
distinguishable from $8$ on this seed set — but it rules out
both endpoints and licenses $8$ as a defensible round-number
default rather than a hand-tuned one.}
\label{tab:abl_full_every_validation}
\resizebox{\linewidth}{!}{%
\begin{tabular}{lccccccccc}
\toprule
$n_{\mathrm{full}}$ & $K\!=\!2$ & $K\!=\!3$ & $K\!=\!4$ & $K\!=\!6$ & $K\!=\!8$ & $K\!=\!12$ & $K\!=\!16$ & $K\!=\!24$ & $K\!=\!32$ \\
\midrule
$1$ ($\equiv$ no\_bd) & $4.245 \pm 0.190$ & $4.072 \pm 0.109$ & $4.049 \pm 0.069$ & $3.485 \pm 0.060$ & $3.005 \pm 0.046$ & $2.506 \pm 0.065$ & $2.316 \pm 0.049$ & $2.208 \pm 0.032$ & $2.187 \pm 0.027$ \\
$2$                   & $2.527 \pm 0.032$ & $2.331 \pm 0.020$ & $2.272 \pm 0.021$ & $2.227 \pm 0.024$ & $2.216 \pm 0.024$ & $2.209 \pm 0.025$ & $2.206 \pm 0.025$ & $2.205 \pm 0.025$ & $2.206 \pm 0.025$ \\
$3$                   & $2.486 \pm 0.037$ & $2.317 \pm 0.023$ & $2.267 \pm 0.024$ & $2.225 \pm 0.023$ & $2.215 \pm 0.024$ & $2.210 \pm 0.024$ & $2.208 \pm 0.023$ & $2.209 \pm 0.024$ & $2.213 \pm 0.024$ \\
$4$                   & $2.466 \pm 0.038$ & $2.308 \pm 0.021$ & $2.262 \pm 0.022$ & $2.222 \pm 0.022$ & $2.213 \pm 0.023$ & $2.209 \pm 0.023$ & $2.208 \pm 0.023$ & $2.211 \pm 0.024$ & $2.215 \pm 0.024$ \\
$6$                   & $2.457 \pm 0.035$ & $2.302 \pm 0.021$ & $2.257 \pm 0.021$ & $2.221 \pm 0.022$ & $2.213 \pm 0.023$ & $2.210 \pm 0.023$ & $2.210 \pm 0.024$ & $2.213 \pm 0.024$ & $2.218 \pm 0.025$ \\
$8$ (reference)       & $2.454 \pm 0.034$ & $2.301 \pm 0.021$ & $2.256 \pm 0.022$ & $2.221 \pm 0.022$ & $2.213 \pm 0.021$ & $2.210 \pm 0.022$ & $2.211 \pm 0.022$ & $2.215 \pm 0.022$ & $2.222 \pm 0.023$ \\
$12$                  & $2.450 \pm 0.038$ & $2.300 \pm 0.024$ & $2.259 \pm 0.026$ & $2.222 \pm 0.025$ & $2.215 \pm 0.025$ & $2.212 \pm 0.025$ & $2.212 \pm 0.025$ & $2.217 \pm 0.025$ & $2.224 \pm 0.025$ \\
$16$                  & $2.447 \pm 0.037$ & $2.300 \pm 0.024$ & $2.257 \pm 0.024$ & $2.222 \pm 0.024$ & $2.214 \pm 0.024$ & $2.211 \pm 0.024$ & $2.212 \pm 0.024$ & $2.218 \pm 0.024$ & $2.226 \pm 0.024$ \\
$24$                  & $2.452 \pm 0.038$ & $2.299 \pm 0.026$ & $2.257 \pm 0.026$ & $2.222 \pm 0.025$ & $2.215 \pm 0.025$ & $2.213 \pm 0.025$ & $2.214 \pm 0.025$ & $2.220 \pm 0.026$ & $2.228 \pm 0.027$ \\
$\infty$              & $2.438 \pm 0.033$ & $2.296 \pm 0.022$ & $2.255 \pm 0.022$ & $2.223 \pm 0.024$ & $2.217 \pm 0.024$ & $2.215 \pm 0.025$ & $2.217 \pm 0.026$ & $2.224 \pm 0.027$ & $2.233 \pm 0.028$ \\
\bottomrule
\end{tabular}
}
\end{table}

\subsection{Sensitivity to the runtime-budget set}
\label{app:runtime_budget_set_sensitivity}

We examine whether it is necessary to sample every integer runtime budget during training. We compare the default ten-point budget set $\mathcal{K}_{\mathrm{Paper\text{-}10}}=\{2,3,4,5,6,8,12,16,24,32\}$ against the dense set $\mathcal{K}_{\mathrm{Dense\text{-}31}}=\{2,3,\ldots,32\}$. Both models contain $287.1$M parameters and are trained for $80.019$M byte-tokens with seed 87. We keep the data, initialization, optimizer, learning-rate schedule, batch configuration, full-budget warmup, and evaluation examples fixed. The mean sampled budgets are also nearly identical ($22.6388$ versus $22.6450$). Both checkpoints are evaluated at every integer budget $K\in\{2,\ldots,32\}$.

\begin{table}[t]
\centering
\small
\caption{Sensitivity to the set of budgets sampled during training. Both checkpoints are evaluated on all 31 integer runtime budgets. Lower BPB is better.}
\label{tab:runtime_budget_set_sensitivity}
\begin{tabular}{lcccc}
\toprule
Training budget set & Mean val.\ BPB & Max.\ gap & $K=2$ & $K=32$ \\
\midrule
Dense-31 & 2.2307 & -- & 2.4838 & 2.2244 \\
Paper-10 & 2.2319 & 0.0030 & 2.4848 & 2.2253 \\
\bottomrule
\end{tabular}
\end{table}

Across all 31 evaluation budgets, the mean absolute validation-BPB difference between Paper-10 and Dense-31 is 0.0012, and the maximum difference is 0.0030 at $K=7$. Paper-10 directly samples only ten budgets during training; among the remaining 21 unseen integer budgets, the maximum linear-interpolation residual is 0.0032 BPB, and no residual exceeds 0.01 BPB. These results indicate that the default ten-point set provides sufficient coverage of the low-, intermediate-, and full-budget regimes in this controlled setting, without requiring every deployable budget to be sampled during training.

This experiment is a single-seed, 287.1M-parameter, 80M-token diagnostic. It supports the adequacy of the default budget set in this setting, but does not establish that the set is minimal or optimal, nor that changing the endpoint range $[2,32]$ would have no effect.

\subsection{Full-budget warmup sensitivity}
\label{app:warmup_sensitivity}

We tested model performance under the default full-budget warmup and under zero
warmup, with all other conditions being the same, in a separate $400.03$ M-byte-token run. Across all
ten standard budgets, the mean and maximum absolute differences are small. We
provide detailed numbers for three representative settings in
\cref{tab:warmup_sensitivity}.

\begin{table}[t]
\centering
\small
\caption{Validation BPB under a 40M-token full-budget warmup and zero warmup.}
\label{tab:warmup_sensitivity}
\begin{tabular}{lccc}
\toprule
Runtime budget $K$ & 40M-token warmup & Zero warmup & Absolute difference \\
\midrule
$2$  & $1.9552$ & $1.9465$ & $0.0087$ \\
$8$  & $1.8221$ & $1.8182$ & $0.0039$ \\
$32$ & $1.8205$ & $1.8204$ & $0.0001$ \\
\bottomrule
\end{tabular}
\end{table}

This provides preliminary evidence that the impact of warmup on the final
result might not be significant. Nevertheless, we retain the warmup as a
precaution during initialization.

\subsection{Training cost at matched full-budget quality}
\label{app:matched_quality_training_cost}

We compare the total training resources required to reach matched full-budget
validation quality. The always-full control defines the target at a validation
BPB of $1.839901$, while the first measured ES-SSM checkpoint that meets this
target achieves a BPB of $1.834882$. As shown in
\cref{tab:matched_quality_training_cost}, ES-SSM processes $17.3\%$ more
byte-tokens and optimization steps, but requires $30.2\%$ less cumulative
analytic compute and $39.4\%$ less measured wall-clock time. Thus, ES-SSM
reaches matched full-budget quality using more updates but less total
computation and substantially less elapsed training time.

\begin{table}[t]
\centering
\small
\caption{Training resources required to reach matched full-budget validation
quality. Resource columns are normalized to the always-full control
($100\%$); lower values indicate lower total cost. The ES-SSM row corresponds
to the first measured checkpoint that meets the target quality.}
\label{tab:matched_quality_training_cost}
\begin{tabular}{lcccc}
\toprule
Method
& \makecell{Full-budget\\val. BPB $\downarrow$}
& \makecell{Steps and\\byte-tokens}
& \makecell{Analytic\\compute}
& \makecell{Measured\\wall-clock} \\
\midrule
Always-full control
& $1.839901$ & $100.0\%$ & $100.0\%$ & $100.0\%$ \\
ES-SSM
& $\mathbf{1.834882}$ & $117.3\%$ & $\mathbf{69.8\%}$ & $\mathbf{60.6\%}$ \\
\bottomrule
\end{tabular}
\end{table}

Analytic compute estimates the number of operations associated with the channels active under each sampled training budget. On a per-update basis, ES-SSM uses
$40.5\%$ less analytic active compute and $48.4\%$ less measured wall-clock
time than always-full training. These per-update savings more than offset the
$17.3\%$ increase in optimization steps. The preceding ES-SSM checkpoint has a
BPB of $1.840758$, while the following checkpoint remains below the target at
$1.837340$, confirming the matched-quality crossing.

%
\section{Complexity and parameter count}
\label{app:complexity}

This appendix makes precise the ``exported compact model is physically smaller''
claim from \cref{sec:Methods}. We give closed-form expressions for the
parameter count and per-token FLOPs of an ES-SSM layer at runtime budget
$K$, identify the fixed and $K$-scalable components, and summarize the
resulting compute and memory trajectories for generic budgets
$K\le\overline K$. All counts below are
\emph{per layer} unless stated otherwise; for an $n_{\mathrm{layers}}$-deep
model, multiply by $n_{\mathrm{layers}}$ and add the shared
embedding/output-head contribution $V\,d$ (tied), where $V$ is the
vocabulary size.

For clarity we assume $d_{\mathrm{in}}=d_{\mathrm{out}}=d$, which is the
setting used in all main-text experiments; the generalization is
immediate.

\subsection{Parameter count at runtime budget $K$}
\label{app:params}

\begin{table}[t]
\centering
\small
\caption{Per-layer parameter ledger. ``Buffer'' entries are cached
spectral quantities and are not learnable parameters.}
\label{tab:param_ledger}
\begin{tabular}{lcc}
\toprule
Component & Expression & Scales with $K$? \\
\midrule
Channel projections $\{M_k\}_{k\le K}$
    & $K\,d^{2}$                          & \textbf{yes} \\
Gate MLP output row $\mathbf{W}_{g,2}[:K,:]$
    & $K\,(d/2)$                          & \textbf{yes} \\
Gate MLP output bias $\mathbf{b}_{g,2}[:K]$
    & $K$                                 & \textbf{yes} \\
\midrule
Gate MLP hidden layer $\mathbf{W}_{g,1},\mathbf{b}_{g,1}$
    & $d\,(d/2) + d/2$                    & no \\
FFN (widening $4$)
    & $8\,d^{2} + 5\,d$                   & no \\
$2\times$ LayerNorm (pre-spectral, pre-FFN)
    & $4\,d$                              & no \\
\midrule
Filter transforms $\{\widehat\Phi_k\}_{k\le K}$ \textit{(buffer)}
    & $K\,(L+1)$ complex                  & (buffer, not counted) \\
\bottomrule
\end{tabular}

\end{table}

Summing the learnable rows of \cref{tab:param_ledger},
\begin{equation}
P_{\mathrm{layer}}(K)
\;=\;
\underbrace{K\big(d^{2}+d/2+1\big)}_{\text{$K$-scalable}}
\;+\;
\underbrace{\tfrac{17}{2}\,d^{2}+\mathrm{O}(d)}_{\text{fixed floor}}.
\label{eq:params_per_layer}
\end{equation}
\Cref{eq:params_per_layer} has two practical consequences. First, the
\emph{marginal} cost of adding one spectral channel is exactly
$d^{2}+d/2+1$ parameters --- an entire $d\times d$ projection plus one
row of the gating MLP. Second, the fixed floor is large: even at
$K=0$ the layer would retain $\tfrac{17}{2}d^{2}+\mathrm{O}(d)$
parameters (gate hidden layer, FFN, LayerNorms). Hence
the prefix-$K$ model is physically smaller than the full model, but
\emph{not} by a factor of $K/\overline K$: the reduction is
$K/\overline K$ only on the channel-scalable part.

Concretely, ignoring $\mathrm{O}(d)$ terms, the ratio of prefix-$K$ to
full parameter counts is
\begin{equation}
\frac{P_{\mathrm{layer}}(K)}{P_{\mathrm{layer}}(\overline K)}
\;=\;
\frac{K+\tfrac{17}{2}}{\overline K+\tfrac{17}{2}}
\quad\xrightarrow[\overline K=32]{}\quad
\frac{K+8.5}{40.5}.
\label{eq:param_ratio}
\end{equation}
At $\overline K=32$, prefix-$K$ models occupy roughly $26\%,\;41\%,\;60\%$ of full-model parameters at $K\!\in\!\{2,8,16\}$ respectively --- a meaningful reduction, but substantially less aggressive than the $K/\overline K$ ratio alone would suggest. We report the exact measured counts in \cref{tab:slimpajama_main}.

\subsection{Per-token FLOPs at runtime budget $K$}
\label{app:flops}

We count multiply--accumulate operations (FLOPs) per input token,
ignoring LayerNorm / activation elementwise work which is subdominant.

\begin{table}[t]
\centering
\small
\caption{Per-token FLOPs ledger. The spectral FFT is amortized across
the sequence; we report per-token cost by dividing the
$\mathrm{O}(Ld\log L)$ sequence-level FFT cost by $L$.}
\label{tab:flops_ledger}
\begin{tabular}{lcc}
\toprule
Stage & FLOPs / token & Scales with $K$? \\
\midrule
Gate MLP hidden               & $d^{2}/2$                    & no \\
Gate MLP output (sliced)      & $K\,d/2$                     & \textbf{yes} \\
Spectral rFFT (input)         & $\mathrm{O}(d\log L)$        & no \\
Spectral complex multiply     & $\mathrm{O}(K\,d)$           & \textbf{yes} \\
Spectral irFFT (output)       & $\mathrm{O}(K\,d\log L)$     & \textbf{yes} \\
Channel projection einsum     & $K\,d^{2}$                   & \textbf{yes} \\
FFN (widening $4$)            & $8\,d^{2}$                   & no \\
\bottomrule
\end{tabular}

\end{table}

Summing the rows of \cref{tab:flops_ledger},
\begin{equation}
C_{\mathrm{layer}}(K)
\;=\;
\underbrace{K\big(d^{2} + \tfrac{d}{2}+\mathrm{O}(d\log L)\big)}_{\text{$K$-scalable}}
\;+\;
\underbrace{\tfrac{17}{2}\,d^{2}+\mathrm{O}(d\log L)}_{\text{fixed floor}}.
\label{eq:flops_per_layer}
\end{equation}
Apart from the $d\log L$ term produced by the FFT, the FLOPs and parameter expressions are structurally identical. The channel projection $Kd^{2}$ dominates at large $K$, while the FFN $8d^{2}$ forms the compute floor at small $K$.

\begin{remark}[Motivation for the $1/\sqrt{K}$ scaling]
\label{rem:sqrt_K}
The factor $1/\sqrt{K}$ is a variance-preserving normalization under
an idealized model of the \emph{gated} channel contributions. Write
$a_k(t)=\operatorname{sigmoid}(s_k(t))M_k(\Phi_k*u)(t)$, so that
$[\mathcal{S}_K(u)](t)=K^{-1/2}\sum_{k=1}^K a_k(t)$.
For a fixed sublayer input distribution, if
$\mathbb{E}a_k(t)=0$, $\mathbb{E}\langle a_i(t),a_j(t)\rangle=0$
for $i\ne j$, and $\mathbb{E}\|a_k(t)\|_2^2\le C_a^2$ uniformly in $k$,
then
\begin{equation}
\mathbb{E}\|[\mathcal{S}_K(u)](t)\|_2^2
=\frac{1}{K}\sum_{k=1}^K\mathbb{E}\|a_k(t)\|_2^2
\le C_a^2.
\label{eq:sqrtK_boundedness}
\end{equation}
This calculation motivates the fixed $1/\sqrt K$ normalization used
across training and inference budgets.
\end{remark}

\section{Experimental setup}
\label{app:setup}

This appendix collects the per-experiment setup details that the main text deferred. Subsections~\ref{app:setup_datasets}--\ref{app:setup_hw} cover datasets, baselines, training protocols, and hardware in turn. Per-experiment hyperparameter tables follow in subsections~\ref{app:setup_slimpajama}--\ref{app:setup_byte_general}.

\subsection{Datasets and tasks}
\label{app:setup_datasets}

\softparagraph{SlimPajama-627B (byte level).} Our main language-modeling benchmark. We stream the SlimPajama-627B corpus~\citep{sobolevaslimpajama} and tokenize raw UTF-8 bytes with a $258$-symbol vocabulary ($256$ bytes plus PAD$=0$ and BOS$=1$). Sequences are sampled at length $L{=}2048$. Validation and test caches contain $1000$ deterministic segments each, built once from a fixed seed and shared bit-for-bit across families.

\softparagraph{Long Range Arena (LRA).} We use six LRA-style long-sequence classification tasks~\citep{tay2020long}: \textsc{ListOps} ($L{=}2{,}000$), \textsc{Text} (IMDb bytes, $L{=}4{,}096$), \textsc{Retrieval} (paired AAN byte sequences, $L{=}4{,}000{\times}2$), \textsc{sCIFAR} (flattened grayscale images, $L{=}1{,}024$), \textsc{Pathfinder} ($L{=}1{,}024$), and \textsc{Path-X} ($L{=}16{,}384$). Reported metric is Top-$1$ accuracy.

\softparagraph{Speech Commands V2 (SC-10).} The ten-class subset of Google Speech Commands V2~\citep{warden2018speech}. Waveforms at $16$\,kHz are converted to $40$-bin log-mel spectrograms with a $25$\,ms window and $10$\,ms hop, yielding $L{=}100$ frames per clip. Reported metric is Top-$1$ accuracy.

\softparagraph{D4RL MuJoCo (offline RL).} The nine canonical D4RL $v2$ datasets~\citep{fu2020d4rl} across three locomotion environments: HalfCheetah, Hopper, Walker2d, each in \textsc{medium}/\textsc{medium-replay}/\textsc{medium-expert} regimes. Returns are normalized with the standard D4RL score, $\mathrm{norm}(G)=(G-G_{\mathrm{rand}})/(G_{\mathrm{expert}}-G_{\mathrm{rand}})\!\times\!100$, using the benchmark's random-policy and expert-policy reference returns for each environment.

\subsection{Models and baselines}
\label{app:setup_models}

We compare ES-SSM against three families of baselines, organized by what each comparison isolates.

\softparagraph{Non-elastic baselines.} \emph{Transformer++} follows the LLaMA recipe~\citep{touvron2023llama}: pre-norm RMSNorm, SwiGLU FFN, RoPE positional encoding, no biases, tied input/output embeddings. \emph{Pythia-style}~\citep{biderman2023pythia} follows the GPT-NeoX/Pythia-style recipe: pre-norm LayerNorm, GELU FFN with widening factor $4$, parallel residual, RoPE applied to $25\%$ of head dimensions, untied embeddings, biases everywhere, with widths and depths resized for our parameter-matched budgets. \emph{GLA}~\citep{yang2023gated} is gated linear attention with low-rank gate projection and SwiGLU FFN. These three families serve as architecture-axis controls: at every reference budget $K\!\in\!\{2,8,16,32\}$ we train an independent fixed-size model whose parameter count matches the corresponding ES-SSM exported compact model within $\pm 5\%$.

\softparagraph{Elastic baselines.} \emph{MatFormer}~\citep{kudugunta2024matformer} is the prior elastic Transformer; we use FFN-width nesting on top of the LLaMA backbone, with nested capacity targets matched to the ES-SSM runtime-budget set reported in \cref{tab:slimpajama_main}. \emph{MatMamba}~\citep{shukla2024matmamba} is the prior elastic SSM with head-dimension nesting on Mamba-2. Both baselines train once under a capacity-sampling schedule over their nested capacity targets, matching the train-once budgeted-inference setting of ES-SSM.

\softparagraph{Static spectral baselines.} Where a fixed-Hankel-basis control is informative, we include a static spectral baseline that removes ES-SSM's input-adaptive gate and budget dropout. On SC-10 this is AR-STU~\citep{agarwal2023spectral}; on LRA this is the non-autoregressive STU baseline in Appendix~\ref{app:add_results_lra}. These rows isolate the contribution of the two ES-SSM mechanisms over a fixed spectral basis.

\softparagraph{Cited rows.} The DT-paper baseline numbers~\citep{chen2021decision} on D4RL MuJoCo are also reported as external references; the paired D4RL comparison uses DT (ours), trained and evaluated under the same protocol as ES-SSM.

\subsection{Training protocols}
\label{app:setup_train}

\softparagraph{Optimizer and schedule (language and classification).} For SlimPajama, LRA, and SC-10, we use AdamW with $(\beta_1,\beta_2){=}(0.9,0.95)$, cosine LR decay after a short linear warmup, global $\ell_2$ gradient clipping at $1.0$, and \texttt{bf16} mixed precision. The ES-SSM spectral branch runs in \texttt{float32} as described in Appendix~\ref{app:precision}. D4RL uses the separate Decision Transformer recipe in Appendix~\ref{app:setup_d4rl}.

\softparagraph{Capacity-sampling schedule (elastic families).} ES-SSM uses budget dropout as defined in \cref{alg:budget_sampler}. MatFormer and MatMamba use analogous sampling over their nested width/head targets. In the SlimPajama scaling experiment, all three elastic families are evaluated on the same ten runtime budgets, with MatFormer / MatMamba capacities indexed by the matched ES-SSM budget in \cref{tab:slimpajama_main}.

\softparagraph{Exported compact models.} After training, each elastic family exports a compact artifact for every evaluated capacity. ES-SSM slices the channel-indexed spectral tensors, including $\{\widehat\Phi_k\}$, $\{M_k\}$, rows of $\mathbf{W}_{g,2}$, and entries of $\mathbf{b}_{g,2}$. MatFormer and MatMamba slice their nested FFN-width or head-dimension tensors. Size, latency, and peak memory are measured on these exported compact models, not on the full-capacity training checkpoint.

\subsection{Hardware}
\label{app:setup_hw}

The main SlimPajama scaling experiment runs on a single NVIDIA B200 ($180$\,GB HBM3e). The byte-level generality replication uses A100 ($80$\,GB) GPUs. Ablations run on A100 ($40$\,GB) cards. The D4RL MuJoCo experiments fit on RTX 3090 ($24$\,GB) cards. All models are implemented in PyTorch $2.4$+. Inference-cost comparisons use the same timing harness, batch size, sequence length, precision policy, and hardware within each benchmark. The ES-SSM spectral branch uses the batched FFT path of Appendix~\ref{app:fft}.

\subsection{SlimPajama scaling: hyperparameters}
\label{app:setup_slimpajama}

\Cref{tab:slimpajama_hp} lists the SlimPajama scaling-experiment hyperparameters, and \cref{tab:slimpajama_esssm_arch} lists the full-budget ES-SSM architecture used for the main run. The measured per-$K$ parameter counts used in the main comparison are reported in \cref{tab:slimpajama_main}.

\begin{table}[t]
\centering
\small
\caption{SlimPajama scaling: training hyperparameters. Identical across all six families except for the capacity schedule: elastic families use a capacity-sampling schedule over their runtime-budget set, while non-elastic families train at a single fixed $K$.}
\label{tab:slimpajama_hp}
\begin{tabular}{ll}
\toprule
Knob & Value \\
\midrule
Sequence length $L$ & $2048$ \\
Vocabulary & $258$ (bytes + PAD + BOS) \\
Token budget & $30$\,B byte-tokens \\
Effective batch & $16$ sequences $\times\,2$ grad accum. $\times\,2048 = 65{,}536$ tokens / step \\
Optimizer & AdamW, $\beta{=}(0.9,0.95)$, $\varepsilon{=}10^{-8}$ \\
Weight decay & $0.1$ on tensor parameters; $0$ otherwise \\
Peak LR & $3\!\times\!10^{-4}$ \\
LR warmup & Linear, first $1\%$ of steps \\
LR decay & Cosine to $10\%$ of peak \\
Gradient clip & Global $\ell_2$-norm $1.0$ \\
Mixed precision & \texttt{bf16} (B200 native), spectral branch in \texttt{fp32} \\
Dropout & $0.1$ \\
Stochastic depth & $p_{\max}=0.1$, linearly scaled across layers \\
ES-SSM budget dropout & \cref{alg:budget_sampler} \\
Budget warmup & First $1$\,B byte-tokens at full budget \\
Validation cadence & Every $1$\,B byte-tokens \\
Hardware & $1\times$ NVIDIA B200 ($180$\,GB) per run \\
\bottomrule
\end{tabular}
\end{table}

\begin{table}[t]
\centering
\small
\caption{SlimPajama main ES-SSM architecture. This is the full-budget
configuration used to train the ES-SSM rows in \cref{tab:slimpajama_main};
smaller ES-SSM entries are exported compact models obtained by slicing this
checkpoint along the spectral-channel axis.}
\label{tab:slimpajama_esssm_arch}
\begin{tabular}{ll}
\toprule
Knob & Value \\
\midrule
$d_{\mathrm{model}}$ & $1262$ \\
$n_{\mathrm{layers}}$ & $20$ \\
Maximum spectral budget $\overline K$ & $32$ \\
Runtime-budget set $\mathcal{K}$ & $\{2,3,4,5,6,8,12,16,24,32\}$ \\
Gate MLP hidden width & $d_{\mathrm{model}}/2=631$ \\
Gate output bias init. & $-2.0$ \\
FFN & GELU MLP, widening factor $4$ \\
Dropout / stochastic depth & $0.1$ / linearly scaled to $0.1$ \\
Embedding / output head & Tied, vocabulary size $258$ \\
Full-budget parameters & around $1.3$\,B \\
\bottomrule
\end{tabular}
\end{table}

\subsection{Long Range Arena: hyperparameters}
\label{app:setup_lra}

The LRA setting is specified in the LRA results appendix rather than repeated
here. Appendix~\ref{app:add_results_lra} gives the matched $\sim\!50$\,M
architecture configurations, baseline roster, dataset variants, per-task
training recipe, and seed policy; see
\cref{tab:lra_arch_configs,tab:lra_train_recipe}.

\subsection{Speech Commands V2: hyperparameters}
\label{app:setup_sc10}

We train on $\sim\!25$\,K-clip SC-10 with the standard $80$/$10$/$10$ split. Each $1$-second clip yields a $L{=}100$ frame log-mel sequence with $40$ mel bins. ES-SSM uses $d{=}256$, $n_{\mathrm{layers}}{=}4$, $\overline K{=}32$. The classification head is a temporal mean pool followed by a single linear projection. Models train for $40$ epochs with batch size $128$, peak LR $3\!\times\!10^{-4}$, $5$ seeds per cell.

\subsection{D4RL: hyperparameters}
\label{app:setup_d4rl}

D4RL~\citep{fu2020d4rl} uses the offline-RL recipe of Decision Transformer~\citep{chen2021decision}: AdamW with $\mathrm{lr}{=}10^{-4}$, weight decay $10^{-4}$, batch size $64$, context length $K_{\mathrm{ctx}}{=}20$, and $150{,}000$ optimization steps. Final evaluation uses $100$ episodes per (model, dataset, seed) at the DT-paper high target return, with three seeds per cell. ES-SSM trains once at $\overline K{=}32$ with the same budget-dropout sampler used elsewhere and exports compact models for the reported budgets; matched Transformer baselines are independently trained at the corresponding parameter budgets.

\subsection{Byte-level generality replication: hyperparameters}
\label{app:setup_byte_general}

We train \emph{Transformer++}, MatFormer, and ES-SSM at around $330$\,M parameters and $10$\,B byte-tokens on the same SlimPajama byte stream. This replication checks that the byte-level data pipeline and full-capacity validation protocol behave consistently before the main SlimPajama scaling study. Tokens-per-step are held at $65{,}536$ for all architectures, with AdamW, cosine decay, gradient clipping, and \texttt{bf16} AMP matched across runs. Validation BPB is logged every $200$\,M tokens on a fixed $1000$-segment cache.

\section{Additional main results}
\label{app:add_results}

This appendix gives the full tables behind the main result figures,
including per-$K$ sweeps, training trajectories, and per-task breakdowns.

\subsection{SlimPajama: full per-$K$ results}
\label{app:slimpajama_full}

\Cref{tab:slimpajama_main} gives the full paired quality--cost table behind
\cref{fig:slimpajama_results}. It covers the complete runtime-budget set
for the elastic families and the independently trained reference points for the
non-elastic baselines.

All ES-SSM rows over $K$ are exported compact models from one trained
$\overline K=32$ checkpoint. Different parameter budgets for
Transformer++, Pythia-style, and GLA are independently trained models, so their
training cost is budget-specific.

\begin{table*}[t]
\centering
\small
\setlength{\tabcolsep}{4pt}
\renewcommand{\arraystretch}{1.05}
\caption{Byte-level SlimPajama main scaling results (1.3B family, 30B byte-token training).
Elastic families (ES-SSM, MatFormer, MatMamba) train once at $\overline K=32$ and are evaluated at all ten runtime budgets in $\mathcal{K}$; non-elastic families (Transformer++, Pythia-style, GLA) contribute four independently trained reference budgets. These non-elastic rows are high-resource reference baselines rather than equal-training-cost competitors, since supporting multiple budgets requires multiple separate training runs. Quality is bits-per-byte (lower is better); inference cost is measured at sequence $L{=}2048$, batch~$1$, on a single B200 GPU.
For ES-SSM, the ten $K$ values instantiate ten exported parameter budgets, spanning roughly $330$\,M to around $1.3$\,B.
A small black downward triangle marks a collapsed test result whose Test BPB is more than $90\%$ higher than the best Test BPB within the same family. All BPB values are rounded to two decimal places.}
\label{tab:slimpajama_main}
\begin{tabular}{lc rrr rrr}
\toprule
& & \multicolumn{3}{c}{\textbf{Quality}} & \multicolumn{3}{c}{\textbf{Inference cost}} \\
\cmidrule(lr){3-5}\cmidrule(lr){6-8}
Family & $K$ & Params (M) & Val BPB & Test BPB & Lat.\ (ms) & Tput (kt/s) & Mem (GB) \\
\midrule
\textbf{ES-SSM (ours)} & 2  &  335.00 & 1.40 & 1.41 &  6.18 & 331.40 & 0.87 \\
\textbf{ES-SSM (ours)} & 3  &  366.90 & 1.22 & 1.23 &  7.50 & 273.10 & 0.93 \\
\textbf{ES-SSM (ours)} & 4  &  398.70 & 1.10 & 1.10 &  8.65 & 236.80 & 1.00 \\
\textbf{ES-SSM (ours)} & 5  &  430.60 & 1.05 & 1.06 &  9.98 & 205.20 & 1.06 \\
\textbf{ES-SSM (ours)} & 6  &  462.50 & 1.02 & 1.03 & 11.34 & 180.60 & 1.13 \\
\textbf{ES-SSM (ours)} & 8  &  526.20 & 0.99 & 0.99 & 14.13 & 144.90 & 1.25 \\
\textbf{ES-SSM (ours)} & 12 &  653.70 & 0.98 & 0.98 & 19.48 & 105.10 & 1.51 \\
\textbf{ES-SSM (ours)} & 16 &  781.10 & 0.96 & 0.97 & 28.60 &  71.60 & 1.76 \\
\textbf{ES-SSM (ours)} & 24 & 1036.00 & 0.95 & 0.95 & 39.42 &  51.90 & 2.27 \\
\textbf{ES-SSM (ours)} & 32 & 1291.00 & 0.95 & 0.96 & 59.83 &  34.20 & 2.78 \\
\midrule
MatFormer & 2  &  341.80 & 5.32 & 5.35 {\scriptsize$\blacktriangledown$} &  4.52 & 453.10 & 0.84 \\
MatFormer & 3  &  379.50 & 5.04 & 5.06 {\scriptsize$\blacktriangledown$} &  5.01 & 408.80 & 0.93 \\
MatFormer & 4  &  419.40 & 4.89 & 4.93 {\scriptsize$\blacktriangledown$} &  5.34 & 383.50 & 1.02 \\
MatFormer & 5  &  460.00 & 4.63 & 4.68 {\scriptsize$\blacktriangledown$} &  5.88 & 348.30 & 1.12 \\
MatFormer & 6  &  502.10 & 4.33 & 4.37 {\scriptsize$\blacktriangledown$} &  6.21 & 329.80 & 1.22 \\
MatFormer & 8  &  577.40 & 3.76 & 3.78 {\scriptsize$\blacktriangledown$} &  7.19 & 284.80 & 1.40 \\
MatFormer & 12 &  718.90 & 3.00 & 3.03 {\scriptsize$\blacktriangledown$} &  8.46 & 242.10 & 1.74 \\
MatFormer & 16 &  855.90 & 2.15 & 2.17 {\scriptsize$\blacktriangledown$} & 10.03 & 204.20 & 2.05 \\
MatFormer & 24 & 1078.60 & 1.24 & 1.27 & 12.03 & 170.20 & 2.59 \\
MatFormer & 32 & 1310.70 & 0.98 & 0.99 & 14.32 & 143.00 & 3.20 \\
\midrule
MatMamba & 2  &  262.00 & 1.63 & 1.65 &  3.02 & 678.10 & 0.58 \\
MatMamba & 3  &  333.00 & 1.29 & 1.30 &  3.71 & 552.00 & 0.74 \\
MatMamba & 4  &  404.00 & 1.10 & 1.12 &  4.34 & 471.90 & 0.89 \\
MatMamba & 5  &  475.00 & 1.06 & 1.08 &  4.99 & 410.40 & 1.05 \\
MatMamba & 6  &  547.00 & 1.04 & 1.05 &  5.81 & 352.50 & 1.21 \\
MatMamba & 8  &  689.00 & 1.02 & 1.04 &  7.27 & 281.70 & 1.52 \\
MatMamba & 12 &  792.00 & 1.01 & 1.02 &  8.09 & 253.20 & 1.75 \\
MatMamba & 16 &  894.00 & 1.00 & 1.01 &  8.81 & 232.50 & 1.97 \\
MatMamba & 24 & 1099.00 & 1.00 & 1.01 & 10.28 & 199.20 & 2.43 \\
MatMamba & 32 & 1310.00 & 1.00 & 1.00 & 11.96 & 171.20 & 2.89 \\
\midrule
Transformer++ & 2  &  354.40 & 1.04 & 1.06 &  5.11 & 400.60 & 0.75 \\
Transformer++ & 8  &  510.30 & 1.00 & 1.01 &  5.74 & 356.90 & 1.03 \\
Transformer++ & 16 &  762.10 & 0.96 & 0.96 &  7.15 & 286.30 & 1.52 \\
Transformer++ & 32 & 1323.80 & 0.93 & 0.94 & 11.49 & 178.30 & 2.63 \\
\midrule
Pythia-style  & 2  &  341.80 & 1.06 & 1.07 &  4.56 & 449.30 & 0.81 \\
Pythia-style  & 8  &  521.30 & 1.01 & 1.02 &  5.64 & 362.90 & 1.15 \\
Pythia-style  & 16 &  798.80 & 0.97 & 0.99 &  6.58 & 311.50 & 1.68 \\
Pythia-style  & 32 & 1286.40 & 0.94 & 0.94 &  9.57 & 214.10 & 2.61 \\
\midrule
GLA           & 2  &  335.00 & 1.07 & 1.08 &  3.94 & 519.80 & 0.62 \\
GLA           & 8  &  526.00 & 1.02 & 1.02 &  4.46 & 459.20 & 0.94 \\
GLA           & 16 &  781.00 & 0.98 & 1.00 &  5.31 & 385.70 & 1.41 \\
GLA           & 32 & 1300.00 & 0.96 & 0.97 &  8.27 & 247.60 & 2.39 \\
\bottomrule
\end{tabular}
\end{table*}



\subsection{Long Range Arena}
\label{app:add_results_lra}

\softparagraph{Setup.} We evaluate the six LRA tasks at the
$\sim\!50$\,M parameter scale. We use $3$ seeds per cell on all six tasks,
including \textsc{Path-X} with its $16{,}384$-token sequence length. ES-SSM trains once at $\overline K{=}32$ and exports
compact models for
$K\!\in\!\mathcal{K}{=}\{2,3,4,5,6,8,12,16,24,32\}$. All four models
are matched at the $\sim\!50$\,M scale with $d_{\text{model}}{=}512$
(\cref{tab:lra_arch_configs}).

\softparagraph{Baselines.} We compare against Transformer with
FlashAttention-2~\citep{dao2023flashattention2},
Mamba-2~\citep{dao2024transformers}, and
STU~\citep{agarwal2023spectral}. STU is the closest mechanism baseline:
it uses the same fixed Hankel top-$K$ basis but lacks the input-adaptive
sigmoid gate and budget-dropout training.

\begin{table}[t]
\centering
\small
\caption{Per-architecture configurations at the matched
$\sim\!50$\,M parameter scale. All models share $d_{\text{model}}{=}512$.
Layer counts are tuned so each backbone stays in the same effective
parameter range; embedding and classifier head are tiny in comparison.}
\label{tab:lra_arch_configs}
\begin{tabular}{lcccc}
\toprule
Model & $d_{\text{model}}$ & $n_{\text{layers}}$ & Specific & Params \\
\midrule
Transformer (FA-2) & $512$ & $16$ & $h{=}8$, RoPE & $50.8$\,M \\
Mamba-2            & $512$ & $13$ & $d_{\text{state}}{=}128$, $\text{headdim}{=}64$ & $50.1$\,M \\
STU                & $512$ & $5$  & $K{=}32$, fixed Hankel basis           & $52.8$\,M \\
\textbf{ES-SSM (ours)} & $512$ & $5$ & $\overline K{=}32$, adaptive gate, budget dropout & $53.5$\,M \\
\bottomrule
\end{tabular}
\end{table}

\softparagraph{Datasets.} We use the standard LRA task definitions:
\textsc{ListOps}, byte-level IMDb \textsc{Text}, AAN \textsc{Retrieval},
\textsc{sCIFAR}, \textsc{Pathfinder}, and \textsc{Path-X}
\citep{maas2011learning,krizhevsky2009learning,linsley2018learning,tay2020long}.
Sequence lengths are listed in \cref{tab:lra_train_recipe};
\textsc{Path-X} is the longest task at $16{,}384$ tokens.

\softparagraph{Training recipe.} All four models use AdamW
($\beta{=}(0.9,0.95)$, weight decay $0.05$, gradient clipping $1.0$),
bf16 mixed precision, gradient checkpointing, and a cosine LR schedule
with learning-rate warmup over the first $10\%$ of optimization steps for
\textsc{ListOps}, \textsc{Text}, \textsc{Retrieval}, and \textsc{Path-X},
and $5\%$ for \textsc{sCIFAR} and \textsc{Pathfinder}. Per-task batch sizes,
training epochs, and peak learning rates are listed in
\cref{tab:lra_train_recipe}.

\softparagraph{LRA budget schedule.} ES-SSM uses no initial full-budget
warmup ($E_{\mathrm{warmup}}=0$). With a zero-based optimization-step counter
$j$, all micro-batches in steps satisfying $j\bmod8=0$ use
$K_{\mathrm{train}}=\overline K=32$. At other steps, each micro-batch draws
its own budget using the log-scale sampler in \cref{eq:budget_sampler}
over $\mathcal{K}=\{2,3,4,5,6,8,12,16,24,32\}$. Thus the full-budget
anchor cadence is $n_{\mathrm{full}}=8$ optimization steps; this schedule
is separate from learning-rate warmup.

\begin{table}[t]
\centering
\small
\caption{Per-task training recipe. Effective global batch is
\texttt{batch} $\times$ \texttt{grad\_accum}.}
\label{tab:lra_train_recipe}
\begin{tabular}{lccccc}
\toprule
Task & Seq.\ len. & Batch & Grad accum & Epochs & LR (peak) \\
\midrule
\textsc{ListOps}    & $2{,}000$  & $32$ & $4$ & $6$  & $3{\times}10^{-4}$ \\
\textsc{Text}       & $4{,}096$  & $16$ & $4$ & $5$  & $2{\times}10^{-4}$ \\
\textsc{Retrieval}  & $4{,}000\,{\times}\,2$ & $8$  & $4$ & $4$  & $2{\times}10^{-4}$ \\
\textsc{sCIFAR} & $1{,}024$  & $64$ & $4$ & $10$ & $4{\times}10^{-4}$ \\
\textsc{Pathfinder} & $1{,}024$  & $64$ & $4$ & $4$  & $3{\times}10^{-4}$ \\
\textsc{Path-X}     & $16{,}384$ & $4$  & $8$ & $2$  & $1{\times}10^{-4}$ \\
\bottomrule
\end{tabular}
\end{table}

\softparagraph{Environment.} We use PyTorch $2.5.1$ with CUDA $12.4$.
Mamba-2 uses its triton SSD kernel, Transformer attention uses PyTorch's
scaled-dot-product attention, and STU / ES-SSM use our implementation.

\softparagraph{Full-capacity comparison.} \cref{tab:lra_full_50m} reports Top-$1$ test accuracy at the full-capacity $\overline K{=}32$ across all six tasks. ES-SSM gives the best result on every task in this matched-scale comparison. The strongest baseline is the static spectral STU model, but ES-SSM still improves it consistently, with the largest margins on \textsc{Text}, \textsc{Retrieval}, and \textsc{Path-X}; these results support the role of the adaptive gate and budget-dropout training beyond a fixed Hankel spectral basis.

\begin{table*}[t]
\centering
\small
\setlength{\tabcolsep}{4pt}
\caption{LRA full-capacity results at the matched
$\sim\!50$\,M parameter scale ($\overline K{=}32$). Top-$1$ test
accuracy (\%), mean$\pm$std over $3$ seeds for every task. Bold marks the per-task best.}
\label{tab:lra_full_50m}
\resizebox{\linewidth}{!}{%
\begin{tabular}{l cccccc}
\toprule
Model & \textsc{ListOps} & \textsc{Text} & \textsc{Retrieval} & \textsc{sCIFAR} & \textsc{Pathfinder} & \textsc{Path-X} \\
\midrule
Transformer (FA-2)     & $48.70{\pm}0.50$ & $76.38{\pm}0.92$ & $70.84{\pm}0.42$ & $61.21{\pm}0.56$ & $84.12{\pm}0.30$ & FAIL \\
Mamba-2                & $49.60{\pm}0.41$ & $85.05{\pm}0.59$ & $81.31{\pm}1.10$ & $78.08{\pm}0.66$ & $81.03{\pm}0.52$ & $74.02{\pm}0.53$ \\
STU                    & $60.10{\pm}0.03$ & $89.25{\pm}1.15$ & $89.68{\pm}0.26$ & $91.24{\pm}0.14$ & $94.28{\pm}0.27$ & $88.01{\pm}0.32$ \\
\midrule
\textbf{ES-SSM (ours)} & $\mathbf{61.70{\pm}0.68}$ & $\mathbf{92.77{\pm}0.60}$ & $\mathbf{91.49{\pm}0.72}$ & $\mathbf{92.88{\pm}0.24}$ & $\mathbf{94.44{\pm}0.20}$ & $\mathbf{93.08{\pm}0.13}$ \\
\bottomrule
\end{tabular}%
}
\end{table*}

\softparagraph{Budgeted inference and sweet spots.}
\Cref{tab:lra_budget_50m,fig:lra_budget_sweep} report ES-SSM's exported compact models
from the same $\overline K{=}32$ training run at all ten budgets in
$\mathcal{K}$ (mean over $3$ seeds for every task). Under
a $98\%$-of-$\overline K$ retention threshold the sweet-spot budget
$K^\star$ varies by task; below the $90\%$-retention collapse
boundary accuracy degrades much faster, consistent with the
budget-dropout training pressure described in \cref{sec:Methods}.
On these completed sweeps, the $98\%$ sweet spots are
$K^\star{=}6$ for \textsc{ListOps}, $6$ for \textsc{Text}, $4$ for
\textsc{Retrieval}, $3$ for \textsc{sCIFAR}, $16$ for
\textsc{Pathfinder}, and $12$ for \textsc{Path-X}. Under the $90\%$
threshold, only \textsc{ListOps} and \textsc{Path-X} cross the collapse
boundary at $K{=}2$; the other tasks remain above it across the evaluated
budget range.

\begin{table}[t]
\centering
\small
\setlength{\tabcolsep}{4pt}
\caption{ES-SSM budget sweep on a single $\overline K{=}32$
training run, $\sim\!50$\,M scale. Test top-$1$ accuracy (\%),
mean over $3$ seeds for every task. \underline{Underlined}
entries mark the smallest $K$ that retains $\ge 98\%$ of the
$\overline K$ score (sweet spot $K^\star$).}
\label{tab:lra_budget_50m}
\begin{tabular}{l cccccccccc}
\toprule
Task & $2$ & $3$ & $4$ & $5$ & $6$ & $8$ & $12$ & $16$ & $24$ & $32$ \\
\midrule
\textsc{ListOps}    & $46.77$ & $56.09$ & $59.73$ & $60.28$ & $\underline{60.77}$ & $61.08$ & $61.33$ & $61.51$ & $61.64$ & $61.70$ \\
\textsc{Text}       & $86.46$ & $87.67$ & $89.52$ & $90.45$ & $\underline{91.10}$ & $91.75$ & $92.21$ & $92.49$ & $92.68$ & $92.77$ \\
\textsc{Retrieval}  & $84.90$ & $88.38$ & $\underline{90.30}$ & $90.48$ & $90.85$ & $91.12$ & $91.31$ & $91.40$ & $91.44$ & $91.49$ \\
\textsc{sCIFAR}     & $89.18$ & $\underline{91.63}$ & $92.17$ & $92.41$ & $92.37$ & $92.55$ & $92.69$ & $92.73$ & $92.81$ & $92.88$ \\
\textsc{Pathfinder} & $85.94$ & $88.58$ & $90.57$ & $91.23$ & $91.61$ & $91.80$ & $92.55$ & $\underline{93.68}$ & $94.16$ & $94.44$ \\
\textsc{Path-X}     & $81.91$ & $85.17$ & $87.50$ & $88.89$ & $89.82$ & $90.75$ & $\underline{91.68}$ & $92.34$ & $92.80$ & $93.08$ \\
\bottomrule
\end{tabular}
\end{table}

\subsection{Speech Commands V2 (audio classification)}
\label{subsec:sc10}
\label{app:add_results_sc10}

\softparagraph{Setup.} SC-10 is the standard $10$-keyword subset of Google Speech Commands V2~\citep{warden2018speech}. We convert each $1$-s waveform to a length-$100$ sequence of $40$-bin log-mel frames and use the official train, validation, and test splits. All sequence models share the same linear audio front-end and mean-pooling classification head, and all reported numbers use the same augmentation and optimizer recipe over $5$ seeds.

\softparagraph{Baselines and protocol.} We compare ES-SSM against Transformer++, Conformer~\citep{gulati2020conformer}, Mamba~\citep{gu2024mamba}, Audio-Mamba~\citep{yadav2024audio}, and AR-STU. All models are matched to within $\pm 5\%$ of $10$\,M trainable parameters. ES-SSM trains once at $\overline K{=}32$ with the same budget dropout schedule used elsewhere, and exports compact models for the ten runtime budgets $\mathcal{K}{=}\{2,3,4,5,6,8,12,16,24,32\}$.

\softparagraph{Results.} \cref{tab:sc10_efficiency} shows that ES-SSM reaches $97.40\%$ full-capacity accuracy, within $0.27$\,pp of Conformer and $0.19$\,pp of the bidirectional Audio-Mamba baseline, while improving the static spectral AR-STU baseline by $2.03$\,pp. This is the mechanism-relevant comparison because AR-STU uses the same Hankel spectral channels but lacks the input-adaptive gate and budget dropout. ES-SSM also matches the best calibration in the table and remains close to AR-STU's FFT-based latency, while running substantially faster than the Conformer and selective-scan baselines.

\softparagraph{Budgeted inference.} \cref{tab:sc10_budget_sweep} reports exported compact ES-SSM$(K)$ classifiers obtained by direct prefix truncation. The main point is that most of the full-capacity audio accuracy is already present at very small budgets: $K{=}3$ retains $99.55\%$ of the $\overline K{=}32$ score, and $K{=}8$ is the best operating point in this sweep. Even $K{=}2$ exceeds the full-capacity AR-STU baseline, showing that the exported compact models are usable well below the full spectral budget.

\begin{table}[t]
\centering
\small
\setlength{\tabcolsep}{5pt}
\caption{SC-10 full-capacity accuracy, inference cost, and
calibration. Top-$1$ accuracy and Expected Calibration
Error~\citep{guo2017calibration} use mean$\pm$std over $5$ seeds. Per-sample
latency is measured at batch~$1$ on an A100 80GB with bfloat16, using the
median over $30$ warmed-up runs. All models are matched at $\sim 10$\,M
parameters. Bold marks the best mean in each column; the ES-SSM row is
the single full-capacity training run from which \cref{tab:sc10_budget_sweep}
is exported.}
\label{tab:sc10_full_budget}
\label{tab:sc10_efficiency}
\begin{tabular}{lcccc}
\toprule
Model & Params & Top-$1$ (\%) & Latency (ms) & ECE \\
\midrule
Conformer                    & $10.31$\,M & $\mathbf{97.67 \pm 0.26}$ & $13.82 \pm 0.19$ & $0.067 \pm 0.014$ \\
Transformer++                & $10.64$\,M & $91.92 \pm 1.71$ & $\phantom{0}4.23 \pm 0.07$ & $0.084 \pm 0.014$ \\
Mamba (causal)               & $10.20$\,M & $97.19 \pm 0.52$ & $\phantom{0}9.15 \pm 0.10$ & $0.086 \pm 0.025$ \\
Audio-Mamba (bidir.)         & $10.20$\,M & $97.59 \pm 0.26$ & $\phantom{0}9.34 \pm 0.18$ & $\mathbf{0.059 \pm 0.005}$ \\
AR-STU                       & $10.51$\,M & $95.37 \pm 0.07$ & $\phantom{0}\mathbf{2.34 \pm 0.04}$ & $0.072 \pm 0.006$ \\
\textbf{ES-SSM (ours, $\overline K{=}32$)} & $10.66$\,M & $97.40 \pm 0.29$ & $\phantom{0}2.97 \pm 0.02$ & $\mathbf{0.059 \pm 0.003}$ \\
\bottomrule
\end{tabular}
\end{table}

\begin{table}[t]
\centering
\small
\setlength{\tabcolsep}{5pt}
\caption{SC-10 budget sweep on a single $\overline K{=}32$
ES-SSM training run. Compact ES-SSM$(K)$ models are exported by
direct prefix truncation of the channel projection stack $\{M_k\}$ and
Hankel spectral channels to the prefix $K$, so the Params column reflects a deployable standalone
model. Top-$1$ test accuracy is mean$\pm$std over $5$ seeds; retention
is relative to $K{=}\overline K{=}32$.
Bold marks the peak accuracy across all $K$.}
\label{tab:sc10_budget_sweep}
\begin{tabular}{cccccc}
\toprule
$K$ & Params & Top-$1$ Acc.\ (\%) & Retention (\%) & Budget (\%) & Latency (ms) \\
\midrule
$\phantom{0}2$ & $2.78$\,M & $96.25 \pm 0.35$           & $\phantom{0}98.81$ & $\phantom{0}6.25$ & $2.95$ \\
$\phantom{0}3$ & $3.04$\,M & $96.96 \pm 0.24$           & $\phantom{0}99.55$ & $\phantom{0}9.38$ & $2.94$ \\
$\phantom{0}4$ & $3.30$\,M & $97.44 \pm 0.28$           & $100.05$           & $12.50$           & $2.97$ \\
$\phantom{0}5$ & $3.56$\,M & $97.61 \pm 0.20$           & $100.22$           & $15.62$           & $2.97$ \\
$\phantom{0}6$ & $3.83$\,M & $97.66 \pm 0.25$           & $100.28$           & $18.75$           & $2.96$ \\
$\phantom{0}8$ & $4.35$\,M & $\mathbf{97.73 \pm 0.15}$  & $\mathbf{100.34}$  & $25.00$           & $2.98$ \\
$12$           & $5.40$\,M & $97.72 \pm 0.10$           & $100.33$           & $37.50$           & $2.95$ \\
$16$           & $6.45$\,M & $97.61 \pm 0.16$           & $100.22$           & $50.00$           & $2.95$ \\
$24$           & $8.56$\,M & $97.52 \pm 0.20$           & $100.13$           & $75.00$           & $3.00$ \\
$32$           & $10.66$\,M & $97.40 \pm 0.29$          & $100.00$           & $100.00$          & $2.95$ \\
\bottomrule
\end{tabular}
\end{table}

\subsection{D4RL offline reinforcement learning}
\label{app:add_results_d4rl}

\softparagraph{Setup.} We replace the Transformer block in Decision Transformer~\citep{chen2021decision} with ES-SSM and keep the standard interleaved $(R_t,s_t,a_t)$ token layout. Models are evaluated by target-return-conditioned rollouts on the nine D4RL MuJoCo datasets~\citep{fu2020d4rl}, using normalized return over $3$ seeds and $100$ episodes per cell. ES-SSM trains once at $\overline K{=}32$ and exports compact models at $K\in\{2,4,8,16,32\}$ by direct prefix truncation; the matched Transformer baselines are separate models trained independently at the corresponding parameter budgets. Therefore, covering the same set of D4RL operating points requires one ES-SSM training run but multiple Transformer training runs, with the latter incurring substantially larger total training time and compute; \cref{tab:d4rl_train_cost} summarizes this accounting.

\begin{table}[t]
\centering
\small
\setlength{\tabcolsep}{4pt}
\caption{D4RL training-cost accounting. A cell denotes one
dataset--seed pair; the full suite contains $9$ datasets and $3$ seeds
($27$ cells). ES-SSM trains one full-capacity model per cell at
$\overline K{=}32$ and exports all five runtime budgets by direct prefix
truncation. The matched Transformer suite trains one independent model
for each budget $K\in\{2,4,8,16,32\}$. Times are single-RTX-3090
training wall-clock; rollout evaluation time is excluded.}
\label{tab:d4rl_train_cost}
\begin{tabular}{lccccc}
\toprule
Protocol & Trainings / cell & Time / training & Suite time / cell & Total time & Relative total \\
\midrule
ES-SSM exported suite
& $1$ & $53.3$ min & $53.3$ min & $23.99$ h & $1.00{\times}$ \\
Transformer 5-$K$ suite
& $5$ & $47.9$ min & $239.6$ min & $107.84$ h & $4.50{\times}$ \\
\midrule
ES-SSM reduction
& $5{\times}$ fewer
& -- 
& $186.3$ min saved
& $83.85$ h saved
& $77.8\%$ lower \\
\bottomrule
\end{tabular}
\end{table}
This accounting is central to the D4RL comparison. A matched Transformer can be competitive at an isolated parameter budget, but covering the full five-budget deployment set requires $5$ independently trained models per dataset--seed cell. ES-SSM obtains the same five operating points from one trained checkpoint, reducing total D4RL training wall-clock from $107.84$ h to $23.99$ h.

\softparagraph{Full-capacity results.} \cref{tab:d4rl_main_full} shows that
ES-SSM at $K{=}32$ improves the average normalized return over our matched DT
baseline by $+4.97$ points. The gain is not uniform across every dataset: it is
driven mainly by \textsc{hopper-medium-replay},
\textsc{walker2d-medium-replay}, and \textsc{hopper-medium-expert}, while the
remaining tasks are close to the DT baseline. The important takeaway is
narrower but useful: the budget-dropout training recipe does not create a
full-capacity penalty on this control benchmark.

\begin{table}[t]
\centering
\small
\setlength{\tabcolsep}{4pt}
\caption{D4RL MuJoCo: normalized returns. Higher is better. Mean $\pm$ std over $3$ seeds, $100$ rollout episodes per cell. ``DT-paper'' is the originally reported number~\citep{chen2021decision}; ``DT (ours)'' is the same architecture trained and evaluated under our recipe ($150$\,k steps, $100$-episode rollout) for paired comparison. ES-SSM at $K{=}32$ matches DT in parameter count ($711$\,k vs.\ $730$\,k). Bold ES-SSM cells: $\mathrm{ES{\text-}SSM}\geq\mathrm{DT (ours)}$. Bold $\Delta$: $|\Delta|>5$.}
\label{tab:d4rl_main_full}
\begin{tabular}{l cccc}
\toprule
Dataset & DT-paper & DT (ours) & \textbf{ES-SSM ($K{=}32$)} & $\Delta$ vs.\ DT (ours) \\
\midrule
halfcheetah-medium        & $42.6$  & $42.22\!\pm\!0.23$  & $\mathbf{42.30\!\pm\!0.09}$ & $+0.08$ \\
halfcheetah-medium-replay & $36.6$  & $34.64\!\pm\!0.80$  & $33.64\!\pm\!2.37$          & $-1.00$ \\
halfcheetah-medium-expert & $86.8$  & $91.12\!\pm\!0.94$  & $\mathbf{91.31\!\pm\!0.77}$ & $+0.19$ \\
hopper-medium             & $67.6$  & $50.91\!\pm\!2.51$  & $\mathbf{52.59\!\pm\!2.74}$ & $+1.68$ \\
hopper-medium-replay      & $82.7$  & $47.96\!\pm\!19.93$ & $\mathbf{79.50\!\pm\!4.02}$ & $\mathbf{+31.54}$ \\
hopper-medium-expert      & $107.6$ & $82.83\!\pm\!17.59$ & $\mathbf{88.06\!\pm\!9.74}$ & $\mathbf{+5.23}$ \\
walker2d-medium           & $74.0$  & $72.86\!\pm\!2.20$  & $71.41\!\pm\!3.04$          & $-1.45$ \\
walker2d-medium-replay    & $66.6$  & $61.04\!\pm\!6.01$  & $\mathbf{69.17\!\pm\!2.50}$ & $\mathbf{+8.13}$ \\
walker2d-medium-expert    & $108.1$ & $103.72\!\pm\!1.14$ & $\mathbf{104.09\!\pm\!8.73}$          & $+0.37$ \\
\midrule
\textbf{Average ($9$ datasets)} & $74.7$ & $65.26$ & $\mathbf{70.23}$ & $+4.97$ \\
\bottomrule
\end{tabular}
\end{table}

\softparagraph{Exported compact models.} \cref{tab:d4rl_infer} compares the exported compact ES-SSM models with separately trained Transformers at matched parameter counts. ES-SSM is ahead for $K\geq 8$, and the smallest two budgets are within $0.15$ normalized-return points of the independently trained Transformers. This makes the deployment claim precise: one ES-SSM training gives several useful operating points on the parameter--quality curve, while matching those points with Transformers requires separate trainings.

\begin{table}[t]
\centering
\small
\setlength{\tabcolsep}{4pt}
\caption{D4RL MuJoCo inference table. ES-SSM rows are exported compact models from a single $\overline K{=}32$ training run; Transformer rows are independently trained baselines (one row $=$ one training). Norm.\ return averaged over $9$ datasets, $3$ seeds, $100$ rollout episodes. Inference cost measured at batch~$1$, $K_{\mathrm{ctx}}{=}20$ on RTX~3090. Bold marks the higher normalized return within each matched budget pair.}
\label{tab:d4rl_infer}
\begin{tabular}{l c rrrrr}
\toprule
Family & $K$ & Params & Norm.\ return & Lat.\ (ms) & Tput (kt/s) & Mem (MB) \\
\midrule
\multicolumn{7}{l}{\emph{ES-SSM exported compact models (one $\overline K{=}32$ training)}} \\
ES-SSM & $32$ & $711$\,k & $\mathbf{70.23}$ & $2.75$ & $21.9$ & $16$ \\
ES-SSM & $16$ & $461$\,k & $\mathbf{65.94}$ & $2.74$ & $21.9$ & $13$ \\
ES-SSM &  $8$ & $335$\,k & $\mathbf{66.55}$ & $2.73$ & $22.0$ & $12$ \\
ES-SSM &  $4$ & $273$\,k & $66.43$ & $2.69$ & $22.3$ & $11$ \\
ES-SSM &  $2$ & $241$\,k & $65.02$ & $2.74$ & $21.9$ & $10$ \\
\midrule
\multicolumn{7}{l}{\emph{Transformer baseline (one independent training per row)}} \\
Transformer (DT) & $32$ & $729$\,k & $65.26$ & $1.61$ & $37.5$ & $12$ \\
Transformer      & $16$ & $436$\,k & $65.73$ & $1.58$ & $38.3$ & $11$ \\
Transformer      &  $8$ & $317$\,k & $64.37$ & $1.58$ & $38.2$ & $11$ \\
Transformer      &  $4$ & $265$\,k & $\mathbf{66.57}$ & $1.56$ & $38.5$ & $10$ \\
Transformer      &  $2$ & $265$\,k & $\mathbf{65.06}$ & $1.55$ & $38.7$ & $10$ \\
\bottomrule
\end{tabular}
\end{table}

\subsection{Byte-level generality replication}
\label{app:add_results_byte_general}

\softparagraph{Setup.} We run a cross-architecture replication of the byte-level training protocol at around $330$\,M parameters / $10$\,B byte-tokens. Three architectures (\emph{Transformer++}, MatFormer, ES-SSM) are trained on the same SlimPajama byte stream with identical optimizer, schedule, effective batch (tokens-per-step $=65{,}536$), seed, validation cache, and gradient-clip; only the architecture differs. The role of this study is to verify that the shared framework reproduces a stable, monotonically converging full-capacity BPB curve for every architecture, and that the relative ordering at full capacity behaves as expected before the main SlimPajama scaling study.

\begin{figure}[t]
\centering
\includegraphics[width=0.92\linewidth]{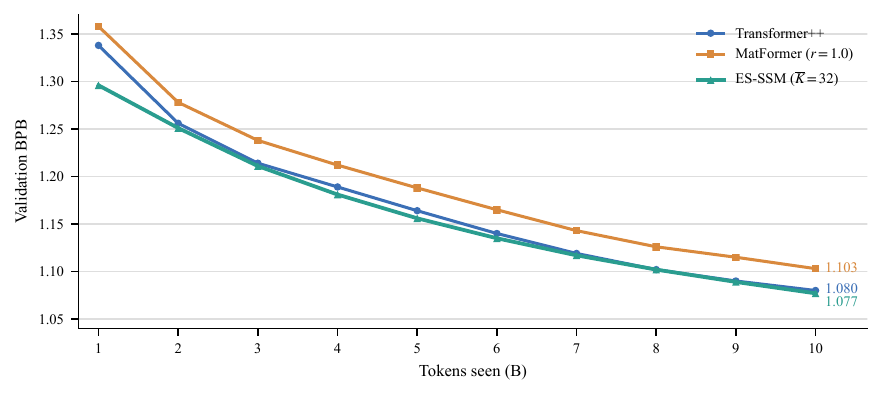}
\caption{Byte-level generality training trajectories. Full-capacity validation
BPB versus tokens seen for the three architectures at the
around $330$\,M-parameter / $10$\,B-byte-token replication scale. Same data as
\cref{tab:reduced_curves}; lower is better.}
\label{fig:reduced_curves}
\end{figure}

\softparagraph{Result.} \Cref{tab:reduced_main} reports the final validation BPB at full capacity ($\overline K{=}32$ for ES-SSM, full FFN width for MatFormer). \Cref{tab:reduced_curves} traces the full-capacity validation BPB at every tenth of the token budget, and \cref{fig:reduced_curves} shows the same trajectories.

\begin{table}[t]
\centering
\small
\setlength{\tabcolsep}{6pt}
\caption{Byte-level generality replication at around $330$\,M parameters and $10$\,B byte-tokens on SlimPajama. Validation BPB on the shared $1000$-segment cache; lower is better. All three runs reach the full $10$\,B-token budget cleanly and write a final \texttt{best.pt} at the last validation gate.}
\label{tab:reduced_main}
\begin{tabular}{l c c c c}
\toprule
Family & Params & Effective batch & \textbf{Final val.\ BPB} ($\downarrow$) & Wall \\
\midrule
Transformer++ (full)                 & $304$\,M & $65{,}536$ tokens/step & $1.0800$ & $\sim\!62$\,h \\
MatFormer (full width)               & $304$\,M & $65{,}536$ tokens/step & $1.1026$          & $\sim\!56$\,h \\
\textbf{ES-SSM ($\overline K{=}32$)} & $335$\,M & $65{,}536$ tokens/step & $\mathbf{1.0771}$          & $\sim\!66$\,h \\
\bottomrule
\end{tabular}
\end{table}

\begin{table}[t]
\centering
\small
\setlength{\tabcolsep}{4pt}
\caption{Full-capacity validation BPB versus tokens seen. Each row is one architecture; columns step the byte-token budget from $1$\,B to $10$\,B in $1$\,B increments. All three trajectories descend monotonically with no observed instability or recovery spike.}
\label{tab:reduced_curves}
\begin{tabular}{l cccccccccc}
\toprule
Tokens (B) & $1$ & $2$ & $3$ & $4$ & $5$ & $6$ & $7$ & $8$ & $9$ & $10$ \\
\midrule
Transformer++              & $1.338$ & $1.256$ & $1.214$ & $1.189$ & $1.164$ & $1.140$ & $1.119$ & $1.102$ & $1.090$ & $1.080$ \\
MatFormer (full width)     & $1.358$ & $1.278$ & $1.238$ & $1.212$ & $1.188$ & $1.165$ & $1.143$ & $1.126$ & $1.115$ & $1.103$ \\
\textbf{ES-SSM ($\overline K{=}32$)} & $1.296$ & $1.251$ & $1.211$ & $1.181$ & $1.156$ & $1.135$ & $1.117$ & $1.102$ & $1.089$ & $\mathbf{1.077}$ \\
\bottomrule
\end{tabular}
\end{table}

\softparagraph{Key observations.} \cref{tab:reduced_main,tab:reduced_curves} show three basic checks.

(\emph{i}) \emph{Matryoshka cost is small.} MatFormer reaches $1.103$ BPB versus $1.080$ for the same-recipe Transformer++, a $+23$\,mBPB cost for supporting four nested FFN ratios.

(\emph{ii}) \emph{ES-SSM is competitive at full capacity.} ES-SSM at $\overline K{=}32$ reaches $1.077$ BPB, essentially tied with Transformer++ at $1.080$ and ahead of MatFormer at $1.103$. Because this replication is single-seed and ES-SSM has a modestly larger parameter count, we read the Transformer++ comparison as parity rather than a large win.

(\emph{iii}) \emph{The shared framework reproduces.} All three validation trajectories descend monotonically, supporting the use of the same byte-level pipeline in the larger study.

\section{Ablations and mechanism diagnostics}
\label{subsec:ablations}
\label{app:ablations}

This section isolates the contribution of each design choice in the ES-SSM spectral sublayer (\cref{subsec:spectral_sublayer}) and the budget dropout training schedule (\cref{subsec:budget_adaptive_training}). The first three subsections are one-variable ablations: each row changes a single knob relative to the reference recipe and holds everything else fixed. The final two subsections are mechanism diagnostics that measure how the trained model uses its spectral channels, rather than the downstream effect of removing one.

\subsection{Protocol}
\label{app:abl_protocol}

All ablations use the same ES-SSM diagnostic preset:
$d\!=\!1024$, $n_{\mathrm{layers}}\!=\!18$,
$\overline K\!=\!32$, sequence length $L\!=\!2048$, and approximately
$770$\,M trainable parameters. Each run trains on SlimPajama for
$400$\,M tokens with AdamW, \texttt{bf16} AMP, the same effective batch,
and seed $86$; remaining hyperparameters follow Appendix~\ref{app:training_recipe}.
Each ablation differs from the reference in exactly one setting.

We evaluate validation BPB at
$K\!\in\!\{2,3,4,5,6,8,12,16,24,32\}$ so that each row shows both
full-capacity and low-budget behavior. Values are next-byte BPB on the
SlimPajama validation cache, taken at the best-val checkpoint. The
ablations are single-seed; pilot runs put the reference variance near
$0.005$ BPB at $K\!=\!\overline K$, and captions flag differences at or
below that noise floor.

\subsection{Budget dropout}
\label{app:abl_bd}

Budget dropout is the training-time mechanism that exposes the model to
multiple runtime budgets. We compare the reference recipe against training
the same architecture only at $K_{\mathrm{train}}\!=\!\overline K$, followed
by direct prefix truncation at evaluation.

\begin{table}[t]
\centering\small
\setlength{\tabcolsep}{4pt}
\renewcommand{\arraystretch}{1.05}
\caption{Without budget dropout, the exported compact model at budget $K$ is a
post hoc direct prefix truncation of a full-capacity model and is not optimised
for any operating point below $\overline K$. At full capacity, the reference
scores $1.826$ BPB versus $1.796$ without budget dropout, a $0.030$ BPB
penalty in this single-seed comparison. In exchange, budget dropout improves
BPB by $0.975$ at $K\!=\!8$, $2.710$ at $K\!=\!3$, and a peak $2.749$
at $K\!=\!4$. These low-budget gains are substantially larger than the
full-capacity penalty and the reported seed noise of this protocol. The
non-monotonicity of the no-budget-dropout row, where $K\!=\!3$ scores worse
than $K\!=\!2$ ($4.583$ vs.\ $3.974\,$BPB) and $K\!=\!4$ even worse still,
makes the same point structurally: without budget dropout, channel ordering
under direct prefix truncation is not a controlled accuracy--compute knob, and
adding one more channel at a small runtime budget can hurt because that channel
was never trained to be used in isolation.}
\label{tab:abl_bd}
\resizebox{\linewidth}{!}{%
\begin{tabular}{lcccccccccc}
\toprule
Training regime & $K\!=\!2$ & $K\!=\!3$ & $K\!=\!4$ & $K\!=\!5$ & $K\!=\!6$ & $K\!=\!8$ & $K\!=\!12$ & $K\!=\!16$ & $K\!=\!24$ & $K\!=\!32$ \\
\midrule
ES-SSM (reference)                         & 1.965 & 1.873 & 1.847 & 1.833 & 1.827 & 1.822 & 1.821 & 1.822 & 1.824 & 1.826 \\
No budget dropout ($\overline K$) & 3.974 & 4.583 & 4.596 & 4.432 & 3.555 & 2.797 & 2.079 & 1.895 & 1.810 & 1.796 \\
\bottomrule
\end{tabular}
}

\end{table}

\subsection{Adaptive channel gating}
\label{app:abl_gate}

The sigmoid gate $g^{(K)}_{k}(t)=K^{-1/2}\operatorname{sigmoid}(s_k(t))$ makes channel contributions input-dependent. Two degenerate limits test whether input adaptivity is the operative property: a uniform gate $g^{(K)}_{k}\!=\!1/\sqrt{K}$, constant across channels and across $t$, and a per-channel scalar gate that learns one trainable weight per channel but remains input-independent.

\begin{table}[t]
\centering\small
\setlength{\tabcolsep}{4pt}
\renewcommand{\arraystretch}{1.05}
\caption{Both degeneracies cost BPB at every $K$. At
$K\!=\!\overline K$ the uniform gate costs $0.226\,$BPB and the
per-channel scalar $0.114\,$BPB relative to the input-adaptive
reference; the corresponding gaps at $K\!=\!2$ are $0.281$ and
$0.139\,$BPB. All four main gaps are an order of magnitude
above the seed noise floor. The per-channel scalar variant sits
between the uniform gate and the input-adaptive reference at
every $K$ and accounts for $0.112$ of the $0.226\,$BPB total
gating gain at $K\!=\!\overline K$ — almost exactly half — suggesting that learned
per-channel reweighting is mildly useful in isolation but that
the dominant share of the gain comes from conditioning on
$u(t)$. The split should be read with the caveat that the
adaptive and per-channel variants differ in capacity (a
two-layer MLP versus $K\!=\!32$ scalars), so the comparison changes both the gate's functional form and its parameter count; a parameter-matched test would require a wider
architectural search beyond this ablation. Direct evidence for
input-conditionality of the reference gate appears in
\cref{tab:diag_gates}, and the order-of-magnitude difference in
effective gate support between adaptive and non-adaptive
variants appears in \cref{tab:diag_keff}.}
\label{tab:abl_gate}
\resizebox{\linewidth}{!}{%
\begin{tabular}{lcccccccccc}
\toprule
Gating variant & $K\!=\!2$ & $K\!=\!3$ & $K\!=\!4$ & $K\!=\!5$ & $K\!=\!6$ & $K\!=\!8$ & $K\!=\!12$ & $K\!=\!16$ & $K\!=\!24$ & $K\!=\!32$ \\
\midrule
Adaptive sigmoid gate (reference)                & 1.965 & 1.873 & 1.847 & 1.833 & 1.827 & 1.822 & 1.821 & 1.822 & 1.824 & 1.826 \\
Uniform $g^{(K)}_{k}\!=\!1/\sqrt{K}$             & 2.246 & 2.131 & 2.091 & 2.068 & 2.058 & 2.049 & 2.043 & 2.041 & 2.045 & 2.052 \\
Per-channel scalar, input-independent            & 2.104 & 2.000 & 1.965 & 1.947 & 1.939 & 1.933 & 1.929 & 1.929 & 1.933 & 1.940 \\
\bottomrule
\end{tabular}%
}

\end{table}

\subsection{Full-budget anchor cadence $n_{\mathrm{full}}$}
\label{app:abl_full_every}

This subsection gives a coarse one-seed ablation of the full-budget
anchor cadence; the finer validation justifying the default is in
Appendix~\ref{app:budget_sampler}. After warmup, forcing
$K_{\mathrm{train}}\!=\!\overline K$ every $n_{\mathrm{full}}$ steps gives
each channel an update-rate floor of $1/n_{\mathrm{full}}$.
\Cref{tab:abl_full_every} checks the endpoints and several
non-endpoint cadences.

\begin{table}[t]
\centering\small
\setlength{\tabcolsep}{4pt}
\renewcommand{\arraystretch}{1.05}
\caption{Coarse full-budget anchor cadence ablation. The endpoint
$n_{\mathrm{full}}\!=\!1$ collapses budget dropout to pure
full-capacity training and shows the same direct-truncation failure mode as
\cref{tab:abl_bd}, as expected. Among the non-endpoint budget-dropout settings
($\infty,16,8,4$), the curve is flat to within $0.02\,$BPB at every
$K$, so this sparse one-seed table should not be used to rank these
cadences. The default $n_{\mathrm{full}}\!=\!8$ is justified by the
finer three-seed validation in Appendix~\ref{app:budget_sampler}; this table
only shows that reasonable non-endpoint cadences preserve the budget
dropout behavior while $n_{\mathrm{full}}\!=\!1$ does not.}
\label{tab:abl_full_every}
\resizebox{\linewidth}{!}{%
\begin{tabular}{lcccccccccc}
\toprule
$n_{\mathrm{full}}$ (steps) & $K\!=\!2$ & $K\!=\!3$ & $K\!=\!4$ & $K\!=\!5$ & $K\!=\!6$ & $K\!=\!8$ & $K\!=\!12$ & $K\!=\!16$ & $K\!=\!24$ & $K\!=\!32$ \\
\midrule
$\infty$ (never force $\overline K$) & 1.955 & 1.869 & 1.845 & 1.832 & 1.826 & 1.823 & 1.823 & 1.825 & 1.829 & 1.834 \\
$16$                                 & 1.960 & 1.871 & 1.847 & 1.834 & 1.828 & 1.825 & 1.824 & 1.825 & 1.828 & 1.831 \\
$8$ (reference)                      & 1.965 & 1.873 & 1.847 & 1.833 & 1.827 & 1.822 & 1.821 & 1.822 & 1.824 & 1.826 \\
$4$                                  & 1.974 & 1.877 & 1.851 & 1.836 & 1.828 & 1.823 & 1.821 & 1.821 & 1.822 & 1.824 \\
$1$ (no budget dropout) & 3.974 & 4.583 & 4.596 & 4.432 & 3.555 & 2.797 & 2.079 & 1.895 & 1.810 & 1.796 \\
\bottomrule
\end{tabular}
}

\end{table}

\bigskip
\noindent The remaining two subsections describe properties
of the trained reference model rather than the effect of a
removed component, anchoring the interpretive claims made
above.

\subsection{Gate statistics}
\label{app:diag_gates}

For a fixed sublayer input, direct prefix truncation preserves the gate value of every retained channel: for any retained channel $k\le K$, $\operatorname{sigmoid}(s_k(t))$ is identical to the value computed in the full-capacity model because slicing away later gate-output rows does not affect row $k$. The only budget-dependent term applied to a retained channel is the deterministic scale $1/\sqrt{K}$. Together with slicing the channel-indexed tensors $\{M_k\}$ and $\{\Phi_k\}$, this lets each exported compact model reuse the same learned prefix without channel search or budget-specific reordering.
In this diagnostic subsection, we write the unscaled gate activity as
$\gamma_k(t)\triangleq\operatorname{sigmoid}(s_k(t))$ and its held-out mean as
$\bar\gamma_k$; this is distinct from the budgeted gate coefficient
$g_k^{(K)}(t)=K^{-1/2}\gamma_k(t)$ used in the model forward pass.

\begin{table}[t]
\centering\small
\setlength{\tabcolsep}{6pt}
\renewcommand{\arraystretch}{1.05}
\caption{Selected-channel sigmoid gate statistics, averaged over
all positions and held-out sequences and across layers. The
trained model assigns substantial gate mass to the low-index channels
($\bar\gamma_k\!\approx\!0.33$ at $k\!=\!2$ and
$\bar\gamma_k\!\approx\!0.15$ at $k\!=\!3$), and the per-channel
standard deviation matches or exceeds the per-channel mean at
every listed channel (most visibly $0.231$ vs.\ $0.150$ at
$k\!=\!3$), so the gate is strongly input-conditional rather
than saturating at a constant. Several high-index channels that
are nearly closed on average ($k\!\ge\!8$,
$\bar\gamma_k\!\le\!0.01$ in our seq-$2048$ measurements) still reach
maximum gate values from about $0.4$ to nearly $1.0$ on specific token contexts; the
gate is opportunistically opening these channels rather than
learning a hard mask. This pattern of input-conditional,
channel-specific use is the load-bearing property
\cref{tab:abl_gate} identifies.}
\label{tab:diag_gates}
\begin{tabular}{lcccc}
\toprule
Channel index $k$ & Mean $\gamma_k$ & Std $\gamma_k$ & Min $\gamma_k$ & Max $\gamma_k$ \\
\midrule
$2$  & $0.326$ & $0.334$ & $0.000$ & $1.000$ \\
$3$  & $0.150$ & $0.231$ & $0.000$ & $0.999$ \\
$4$  & $0.068$ & $0.142$ & $0.000$ & $0.995$ \\
$5$  & $0.032$ & $0.091$ & $0.000$ & $0.987$ \\
$6$  & $0.024$ & $0.061$ & $0.000$ & $0.969$ \\
$8$  & $0.007$ & $0.029$ & $0.000$ & $0.957$ \\
$12$ & $0.001$ & $0.001$ & $0.000$ & $0.075$ \\
$16$ & $0.001$ & $0.004$ & $0.000$ & $0.443$ \\
$24$ & $0.001$ & $0.004$ & $0.000$ & $0.665$ \\
$32$ & $0.001$ & $0.007$ & $0.000$ & $0.550$ \\
\bottomrule
\end{tabular}

\end{table}

\subsection{Effective gate support}
\label{app:diag_keff}

We quantify how concentrated the gate activity is across spectral channels with two complementary measures. Both are computed from the per-channel mean unscaled gate activities $\bar\gamma_k$ measured on the same held-out cache as \cref{tab:diag_gates}, using all spectral channels for each gate variant:
\begin{align}
K_{\mathrm{eff}}^{\,\mathrm{PR}}
  & \;\triangleq\; \big(\textstyle\sum_k \bar\gamma_k\big)^{\!2}
  \,\Big/\, \textstyle\sum_k \bar\gamma_k^{\,2}, \tag{participation ratio}\\
K_{\mathrm{eff}}^{\,(\tau)}
  & \;\triangleq\; \#\{k : \bar\gamma_k \ge \tau\}. \tag{threshold count at $\tau$}
\end{align}
The participation ratio equals the number of channels included in the sum when the gate distribution is uniform and $1$ when a single channel dominates; $K_{\mathrm{eff}}^{\,(0.05)}$ counts channels whose mean unscaled gate activity is at least $0.05$ on the held-out cache -- a coarser but more interpretable proxy.

\begin{table}[t]
\centering\small
\setlength{\tabcolsep}{6pt}
\renewcommand{\arraystretch}{1.05}
\caption{Effective gate support for the gate variants, computed
from the per-channel mean unscaled gate activities measured on the held-out
validation cache. The adaptive gate concentrates gate activity into a
small number of channels: the reference model has participation ratio
$3.84$, and only four channels exceed the $0.05$ mean-activity threshold.
Removing budget dropout changes this count only modestly, while the
non-adaptive variants fail to sparsify: the uniform gate uses all
channels by construction, and the per-channel scalar remains broadly
spread across the channel set. This supports the interpretation that
input-adaptive gating, not merely a static channel weight, concentrates
gate activity onto a compact subset of channels. This diagnostic measures
gate activity only, rather than the full channel contribution
$g_k^{(K)}(t)M_k(\Phi_k*u)(t)$; low mean-activity channels may still provide
input-specific high-budget refinements.}
\label{tab:diag_keff}
\begin{tabular}{lccc}
\toprule
Variant
   & $K_{\mathrm{eff}}^{\,\mathrm{PR}}$
   & $K_{\mathrm{eff}}^{\,(0.05)}$
   & $K_{\mathrm{eff}}^{\,(0.01)}$ \\
\midrule
Adaptive sigmoid gate (reference)             & $\phantom{0}3.84$  & $\phantom{0}4$ & $\phantom{0}7$ \\
Adaptive, no budget dropout (\textsc{no\_bd}) & $\phantom{0}4.35$  & $\phantom{0}4$ & $\phantom{0}7$ \\
Per-channel scalar  (\textsc{gate\_static})   & $28.25$            & $32$           & $32$ \\
Uniform $1/\sqrt{K}$  (\textsc{gate\_uniform})  & $32.00$            & $32$           & $32$ \\
\bottomrule
\end{tabular}

\end{table}


\end{document}